\documentclass[twoside,11pt]{article}
\usepackage{jmlr2e}
\usepackage{graphicx}
\usepackage{url}
\usepackage{booktabs}
\usepackage{amsmath}
\usepackage{amssymb}
\usepackage{bbm}
\usepackage{enumerate}
\usepackage{tabularx}
\usepackage[table]{xcolor}
\definecolor{tableShade}{gray}{0.9}
\definecolor{lightblue}{RGB}{230,247,254}
\definecolor{lightgreen}{RGB}{222,251,152}
\definecolor{lightorange}{RGB}{254,239,224}
\usepackage{colortbl}
  \usepackage{multirow}
  \usepackage{graphicx}
  \usepackage{soul,color,url}  

  \usepackage{bbding}
  \usepackage{colortbl}
  \usepackage{bm}
  \usepackage{subfigure,multirow}
  \usepackage{listings}             
  \usepackage{wrapfig}
  \usepackage{framed}
\definecolor{shadecolor}{RGB}{230,247,254} 

  \newenvironment{SnugshadeF}[1][254,239,224]{
   \definecolor{shadecolor}{RGB}{#1}%
  \begin{snugshade}%
}{%
    \end{snugshade}%
}

  \newenvironment{SnugshadeB}[1][230,247,254]{
  \begin{snugshade}%
}{%
    \end{snugshade}%
}

\newcounter{MethodF}
\newenvironment{MethodF}[1][]{\refstepcounter{MethodF}\par\noindent\textbf{Frequentist #1\\}\begin{upshape}}
{\end{upshape} \par}

  \newcounter{MethodB}
\newenvironment{MethodB}[1][]{\refstepcounter{MethodB}\par\noindent\textbf{Bayesian #1\\}\begin{upshape}}
{\end{upshape} \par}

\begin{document}

   \newcommand{\xmark}{\ding{55}}%
   
  \title{Time for a Change: a Tutorial for Comparing Multiple Classifiers Through Bayesian Analysis}

\author{\name Alessio Benavoli$^\dagger$\email alessio@idsia.ch \\
\name Giorgio Corani$^\dagger$\email giorgio@idsia.ch \\
\name Janez Dem\v{s}ar$^\natural$ \email janez.demsar@fri.uni-lj.si\\
\name Marco Zaffalon$^\dagger$\email zaffalon@idsia.ch \\
\addr $^\natural$Faculty of Computer and Information Science, University of Ljubljana,\\
Vecna pot 113, SI-1000 Ljubljana,  Slovenia\\
       \addr $^\dagger$Istituto Dalle Molle di Studi sull'Intelligenza Artificiale (IDSIA)\\
          Galleria 2, 6928 Manno, Switzerland\\
}
  \editor{} 
  




  
  \maketitle
  
  \begin{abstract}
The machine learning community adopted the use of null hypothesis significance testing (NHST) in order to ensure the statistical validity of results. Many scientific fields however realized the shortcomings of frequentist reasoning and in the most radical cases even banned its use in publications. We should do the same: just as we have embraced the Bayesian paradigm in the development of new machine learning methods, so we should also use it in the analysis of our own results. We argue for abandonment of NHST by exposing its fallacies and, more importantly, offer better---more sound and useful---alternatives for it.
\end{abstract}

  
  \section{Introduction}
Progression of Science and of the scientific method go hand in hand. Development of new theories requires---and at the same time facilitates---development of new methods for their validation.

Pioneers of machine learning were playing with ideas: new approaches, such as induction of classification trees, were worthy of publication for the sake of their interestingness. As the field progressed and found more practical uses, variations of similar ideas began emerging, and with that the interest in determining which of them work better in practice. A typical example are the different measures for assessing the quality of attributes; deciding which work better than others required tests on actual, real-world data. Papers thus kept introducing new methods and measured, for instance, classification accuracies to prove their advantages over the existing methods. To ensure the validity of such claims, we adopted---starting with the work of~\cite{Dietterich98} and~\cite{Salzberg97}, and later followed by~\cite{demvsar2006statistical}---the common statistical methodology used in all scientific areas relying on empirical observations: the null hypothesis significance testing (NHST).

This spread the understanding that the observed results require statistical validation. On the other hand, NHST soon proved inadequate for many reasons~\citep{demvsar2008appropriateness}.
Noteworthy, the American Statistical Association has recently made a statement against p-values   \citep{wasserstein2016asa}. NHST nowadays it is also falling out of favour in other fields of science~\citep{banned}.  We believe that the field of machine learning is ripe for a change as well.

We will spend a whole section demonstrating the many problems of NHST. In a nutshell: it does not answer the question we ask. In a typical scenario, a researcher proposes a new method and desires to prove that it is more accurate than another method on a \textit{single} data set or on a \textit{collection} of data sets. She thus runs the competing methods and records their results (classification accuracy or another appropriate score) on one or more data sets, which is followed by NHST. The difference between what the researcher has in mind and what the NHST provides for is evident from the following quote from a recently published paper: 
``{\em Therefore, at the 90\% confidence level, we can conclude that (...) method is able to significantly outperform the other approaches}.''  This is wrong. The stated {\em 90\% confidence level} is not the probability of one classifier outperforming another. The NHST computes the probability of getting the observed (or a larger) difference between classifiers if the null hypothesis of equivalence was true, which is not the probability of one classifier being more accurate than another, given the observed empirical results. Another common problem is that the claimed statistical significance might have no practical impact. Indeed, the common usage of NHST relies on the wrong assumptions that the $p$-value is a reasonable proxy for the probability of the null hypothesis and that statistical significance implies practical significance. 

As we wrote at the beginning, development of Science not only requires but also facilitates the improvement of scientific methods. Advancement of computational techniques and power reinvigorated the interest for Bayesian statistics. Bayesian modelling is now widely adopted for designing principled algorithms for learning from data \citep{bishop2007pattern,murphy2012machine}. It is time to also switch to Bayesian statistics when it comes to analysis of our own results.

The questions we are actually interested in---e.g., is method A better than B? Based on the experiments, how probably is A better? How high is the probability that A is better by more than 1\%?---are questions about posterior probabilities. These are naturally provided by the Bayesian methods \citep{edwards1963bayesian,dickey1973scientific,berger1987testing}. The core of this paper is thus a section that establishes the Bayesian alternatives to frequentist NHST and discusses their inference and results. We eventually describe also the software libraries with the necessary algorithms and give short instructions for their use.

  \section{Frequentist analysis of experimental results}
  \label{s-freq}
    Why do we need to go beyond the frequentist  analysis of experimental results? To answer this question,
 we will focus on a practical case: the comparison of the accuracy of classifiers on different datasets.
  We initially consider two classifiers: naive Bayes (nbc) and averaged one-dependence estimator (aode).
 A description of these algorithms with exhaustive references is given in the book by \cite{witten2011data}.
Assume that our aim is to compare \textit{nbc} versus \textit{aode}.
  These are the steps we must follow:
  \begin{enumerate}
  \item choose a comparison metric;
   \item select a group of datasets to evaluate the algorithms; 
   \item perform $m$ runs of $k$-fold cross-validation for each classifier on each dataset.
  \end{enumerate}
We have performed these steps in WEKA, choosing \textit{accuracy} as metric, on a collection of $54$ data sets  downloaded from the the WEKA website\footnote{See \url{http://www.cs.waikato.ac.nz/ml/weka/}.}  and with $10$ runs of $10$-fold cross-validation.
 Table \ref{tab:accuracies} reports the accuracies obtained on each dataset by each classifier.
 
 First of all, we aim at knowing which is the best classifier for each  dataset. 
 The answer to this question is probabilistic, since on each data set we have only estimates 
 of the performance of each classifier.
 
 \begin{table}[h]
    \rowcolors{1}{white}{lightblue}
  \begin{center}
  \begin{tabular}{rrrrrrrr}
\text{Datasets}  & \multicolumn{7}{c}{\text{10 runs of 10-fold cross-validation}}\\\hline
anneal & 94.44  & 98.89  & 94.44  & 98.89 & \dots & 94.38 & 97.75\\
anneal & 96.67  & 100.0  & 96.67  & 100.0 & \dots & 96.63 & 97.75\\
audiology & 73.91  & 69.56  & 73.91  & 60.87 & \dots & 72.73 & 59.09\\
audiology & 73.91  & 69.56  & 78.26  & 60.87 & \dots & 72.73 & 59.09\\
breast-cancer & 90.32  & 90.32  & 87.1  & 86.67 & \dots & 86.67 & 90.0\\
breast-cancer & 87.1  & 87.1  & 87.1  & 86.67 & \dots & 83.33 & 86.67\\
cmc & 51.35  & 50.68  & 54.73  & 59.18 & \dots & 50.34 & 48.3\\
cmc & 52.7  & 50.68  & 52.7  & 55.1 & \dots & 52.38 & 48.98\\
\dots & \dots & \dots& \dots & \dots & \dots & \dots & \dots \\
\dots & \dots & \dots& \dots & \dots & \dots & \dots & \dots \\
wine & 100.0  & 95.71  & 97.14  & 94.29 & \dots & 97.14 & 97.1\\
wine & 100.0  & 95.71  & 97.14  & 92.86 & \dots & 97.14 & 97.1\\
yeast & 57.72  & 55.03  & 59.06  & 58.39 & \dots & 55.4 & 55.4\\
yeast & 57.05  & 55.03  & 59.06  & 58.39 & \dots & 54.05 & 55.4\\
zoo & 81.82  & 100.0  & 100.0  & 90.0 & \dots & 90.0 & 100.0\\
zoo & 90.91  & 100.0  & 100.0  & 90.0 & \dots & 90.0 & 100.0\\
  \hline
  \end{tabular}
  \end{center}
  \caption{Accuracies for 10 runs of 10-fold cross-validation on $54$ UCI datasets for \textit{nbc} (blue row)
  versus \textit{aode} (white rows).}.
  \label{tab:accuracies}
  \end{table}
  
Since during cross-validation we have provided both classifiers with the same  training and test sets, 
we can compare the classifiers by considering the difference in accuracies  on each test set.
This yields the vector of \textit{differences of accuracies} $\bm{x}=\{x_1,x_2,\dots,x_n\}$, where $n=100$ ($10$ runs of $10$-fold cross-validation). 
We can compute the mean of  the vector of differences $\bm{x}$, i.e., the mean difference of accuracy between the two classifiers,  and statistically evaluate whether the mean difference is significantly different from zero.
In frequentist analysis, we would perform a NHST using  the  $t$-test.
The problem is that the $t$-test assumes the observations to be independent.
However, the differences of accuracies are not independent of each other because of the overlapping training sets used in cross-validation.
  Thus the  usual $t$-test is \textit{not} calibrated when applied to the analysis of cross-validation
  results:  when sampling the data under the null hypothesis its rate of Type I errors is much larger than  $\alpha$ \citep{Dietterich98}.
  Moreover, the correlation cannot be estimated from data; \cite{nadeau2003inference} have proven that there is no unbiased estimator of the correlation of the results obtained on the different folds.  Introducing some approximations, they have proposed a heuristic to choose the correlation  parameter: $\rho=\frac{n_{te}}{n_{tot}}$, where
  $n_{te}$, $n_{tr}$ and $n_{tot}=n_{te}+n_{tr}$ respectively denote the size of the training set, of the test set and of the whole available data set.\footnote{\cite{nadeau2003inference} considered the case in which random training and test sets are drawn from the original data set.   This is slightly different from k-fold cross-validation, in which the folds are designed not to overlap. However  the correlation heuristic by \cite{nadeau2003inference} has since become  commonly used to analyse the cross-validation results \citep{bouckaert2003choosing}.}

  \begin{SnugshadeF}
\begin{MethodF}[correlated t-test]
The correlated $t$-test is based on the modified Student's t-statistic:
  \begin{equation}
  t(\bm{x},\mu)=
  \frac{\overline{x}-\mu}{\sqrt{\hat{\sigma}^2(\frac{1}{n}+\frac{\rho}{1-\rho})}}=
  \frac{\overline{x}-\mu}{\sqrt{\hat{\sigma}^2(\frac{1}{n}+\frac{n_{te}}{n_{tr}})}},\label{eq:correlated-stat} 
  \end{equation}
  where $\overline{x}=\frac{1}{n}\sum_{i=1}^n x_i$ and $\hat{\sigma}=\sqrt{\frac{1}{n-1} \sum_{i=1}^n (x_i-\overline{x})^2}$
  are the sample mean and sample standard deviation of the data $\bm{x}$, $\rho$   is the correlation between the observations  and $\mu$ is the value of the mean we aim at testing.   The statistic follows a Student distribution with $n-1$ degrees of freedom:
      \begin{equation}
\label{eq:freqstud}
      St\left(\bar{x};n-1,\mu,\left(\frac{1}{n}+\frac{\rho}{1-\rho}\right)\hat{\sigma}^2\right).
  \end{equation}
  For $\rho=0$, we obtain the traditional  $t$-test.
  For $\rho=\frac{n_{te}}{n_{tot}}$, we obtain the correlated $t$-test proposed by  \cite{nadeau2003inference}
  to account for the correlation due to the overlapping training sets.
Usually the test is run in a two-sided fashion. Its hypotheses are:
  $H_0: \mu = 0; \,\,\,\,\, H_1: \mu \neq 0$.
  The p-value of the statistic under the null hypotheses is:
  \begin{equation} 
  p=2\cdot(1-\mathcal{T}_{n-1}(|t(\bm{x},0)|)),
 \end{equation}
  where $\mathcal{T}_{n-1}(|t(\bm{x},0)|)$ denotes the cumulative distribution of the standardized Student distribution with $n-1$ degrees of freedom
  in $|t(\bm{x},\mu)|$ for $\mu=0$.     For instance, for the first data set in Table \ref{tab:accuracies} we have that $\bar{x}=-0.0194$, $\hat{\sigma}=0.01583$, $\rho=1/10$, $n=100$ and so
  $t(\bar{x},0)=-3.52$. Hence, the two-sided $p$-value is $p=2\cdot(1-\mathcal{T}_{n-1}(|t(\bm{x},0)|))=0.00065\approx 0.001$.  
  Sometimes the directional one-sided test is performed. 
  If the alternative hypothesis is the positive one, the hypotheses of the one-sided test are: $H_0: \mu \leq 0; \,\,\,\,\, H_1: \mu > 0$.
  The p-value is $ p=1-\mathcal{T}_{n-1}(t(\bm{x},0))$.
\end{MethodF}
\end{SnugshadeF}

Table \ref{tab:pvalues} reports the two-sided $p$-values for each comparison on each dataset obtained via the correlated t-test. The common practice 
in NHST is to declare all the comparisons such that $p\le 0.05$ as significant, i.e., the accuracy of the two classifiers is significantly
different on that dataset. Conversely, all the comparisons with $p> 0.05$ are declared  not significant.
Note that these significance tests can, under the NHST paradigm, only be considered in isolation, while combined they require either an omnibus test like ANOVA or corrections for multiple comparisons.

  \begin{table}[h]
    \rowcolors{1}{white}{lightorange}
  \begin{center}
  \begin{tabular}{rrrrrr}
\text{Dataset}  & \multicolumn{1}{c}{$p$-value} & \text{Dataset}  & \multicolumn{1}{c}{$p$-value} & \text{Dataset}  & \multicolumn{1}{c}{$p$-value}\\\hline
anneal & 0.001 & audiology & 0.622 & breast-cancer & 0.598 \\
cmc & 0.338 & contact-lenses & 0.643 & credit & 0.479 \\
german-credit & 0.171 & pima-diabetes & 0.781 & ecoli & 0.001 \\
eucalyptus & 0.258 & glass & 0.162 & grub-damage & 0.090 \\
haberman & 0.671 & hayes-roth & 1.000 & cleeland-14 & 0.525 \\
hungarian-14 & 0.878 & hepatitis & 0.048 & hypothyroid & 0.287 \\
ionosphere & 0.684 & iris & 0.000 & kr-s-kp & 0.646 \\
labor & 1.000 & lier-disorders & 0.270 & lymphography & 0.018 \\
monks1 & 0.000 & monks3 & 0.220 & monks & 0.000 \\
mushroom & 0.000 & nursery & 0.000 & optdigits & 0.000 \\
page & 0.687 & pasture & 0.000 & pendigits & 0.452 \\
postoperatie & 0.582 & primary-tumor & 0.492 & segment & 0.000 \\
solar-flare-C & 0.035 & solar-flare-m & 0.596 & solar-flare-X & 0.004 \\
sonar & 0.777 & soybean & 0.049 & spambase & 0.000 \\
spect-reordered & 0.198 & splice & 0.004 & squash-stored & 0.940 \\
squash-unstored & 0.304 & tae & 0.684 & credit & 0.000 \\
owel & 0.000 & waveform & 0.417 & white-clover & 0.463 \\
wine & 0.671 & yeast & 0.576 & zoo & 0.435 \\
  \hline 
  \end{tabular}
  \end{center}
  \caption{Two sided $p$-values for each dataset. The difference is significant ($p<0.05$) in 19 out of 54 comparisons.}
  \label{tab:pvalues}
  \end{table}

\subsection{NHST: the pitfalls of black and white thinking}
  Despite being criticized from its inception, NHST is still considered necessary for 
  publication, as $p\le 0.05$ is trusted as an objective proof of the method's quality.
  One of the key problems of decisions based on $p\le 0.05$ is that it leads to ``black and white thinking'',
which ignores the fact that (i) a \textit{statistically significant difference} is completely different from a \textit{practically significant difference} \citep{berger1987testing};
  (ii) two methods that \textit{are not statistically significantly different} \textit{are not necessarily equivalent}.
  The NHST and this $p$-value-related ``black and white thinking'' do not allow for making informed decisions.
  Hereafter, we list the limits of NHST in order of severity using, as a working example, the assessment of the performance of classifiers.

  \paragraph{NHST does not estimate probabilities of hypotheses.} {\em What is the  probability that the performance of two classifiers is different (or equal)?}
  This is \textit{the question} we are asking when we compare two classifiers; and NHST cannot answer it.
  
  In fact, the $p$-value represents the probability of getting the observed (or larger) differences assuming that the performance of the classifiers is equal ($H_0$). Formally, $p=p(t(\bm{x})>\tau|H_0)$, where $t(\bm{x})$ is the statistic computed from the data $\bm{x}$, and $\tau$ is the critical value corresponding to the test and the selected $\alpha$. This is not the probability of the hypothesis, $p(H_0|\bm{x})$, in which we are interested.

  Yet, researchers want to know the probability of the null and the alternative hypotheses on the basis of the observed data, rather than the probability of the data assuming the null hypothesis to be true. Sentences like ``{\em at the 95\% confidence level, we can conclude that }(...)'', are formally correct, but they seem to imply that $1-p$ is the probability 
of the alternative hypothesis, while in fact $1 - p = 1 - p(t(\bm{x})>\tau|H_0) = p(t(\bm{x})<\tau|H_0)$, which is not the same as $p(H_1|\bm{x})$. 
  This is summed up in Table \ref{tab:nhstwrong}.
  
  \begin{table}[h]
\newcommand{\mc}[2]{\multicolumn{#1}{c}{#2}}
\newcolumntype{a}{>{\columncolor{lightorange}}c}
\newcolumntype{b}{>{\columncolor{lightblue}}c}
  \centering
     \begin{tabular}{ ab }
   \multicolumn{1}{c}{\textbf{what we compute}}
 & \multicolumn{1}{c}{\textbf{what we would like to know}} \\
\midrule
 $p(t(\bm{x})>\tau|H_0)$ & $p(H_0|\bm{x})$\\
$1-p(t(\bm{x})>\tau|H_0)=p(t(\bm{x})<\tau|H_0)$ & $1-p(H_0|\bm{x})=p(H_1|\bm{x})$\\
\bottomrule
\end{tabular}
\caption{Difference between the probabilities of interest for the analyst and the probabilities computed by the frequentist test.}
\label{tab:nhstwrong}
  \end{table}




    \paragraph{Point-wise null hypotheses are practically always false.}   \textit{The difference between two classifiers can be very small; however there are no two classifiers whose 
  accuracies are perfectly equivalent.} 

By using a NHST, the null hypothesis is that the classifiers are equal. However, the null hypothesis is practically always false!
 By rejecting the null hypothesis NHST indicates that the null hypothesis is unlikely; but this is known even before running the experiment.
  This problem of the NHST has been pointed out in many different scientific domains
  \cite[Sec 4.1.2.2]{lecoutre2014significance}.
  A consequence is that, since the null hypothesis is always false, 
  \textbf{by adding enough data points it is possible to claim significance even when the effect size is trivial}.
  This is because the p-value is affected both by the sample size and the effect size, as discussed in the next section. Quoting \cite{kruschke2015bayesian}: \textit{``null hypotheses    are straw men that can virtually always be rejected with
  enough data.''}

\paragraph{The $p$-value does not separate between the effect size and the sample size.} The usual explanation for this phenomenon is that if the effect size $H_0$ is small, more data is needed to demonstrate the difference. Enough data can confirm arbitrarily small effects. Since the sample size is manipulated by the researcher and the null hypothesis is always wrong, \textit{the researcher can reject it by testing the classifiers on enough data.} 
   On the contrary, conceivable differences may fail to yield small $p$-values if there are not enough suitable data for testing the method (e.g., not enough datasets).
     Even if we pay attention not to confuse the $p$-value with the probability of the null hypothesis, the $p$-value is intuitively understood as the indicator of the effect size. In practice, it is the function of effect size and sample size: same $p$-values do not imply same effect sizes.
     
\begin{figure}[h]
    \centering
        \includegraphics[width=7cm]{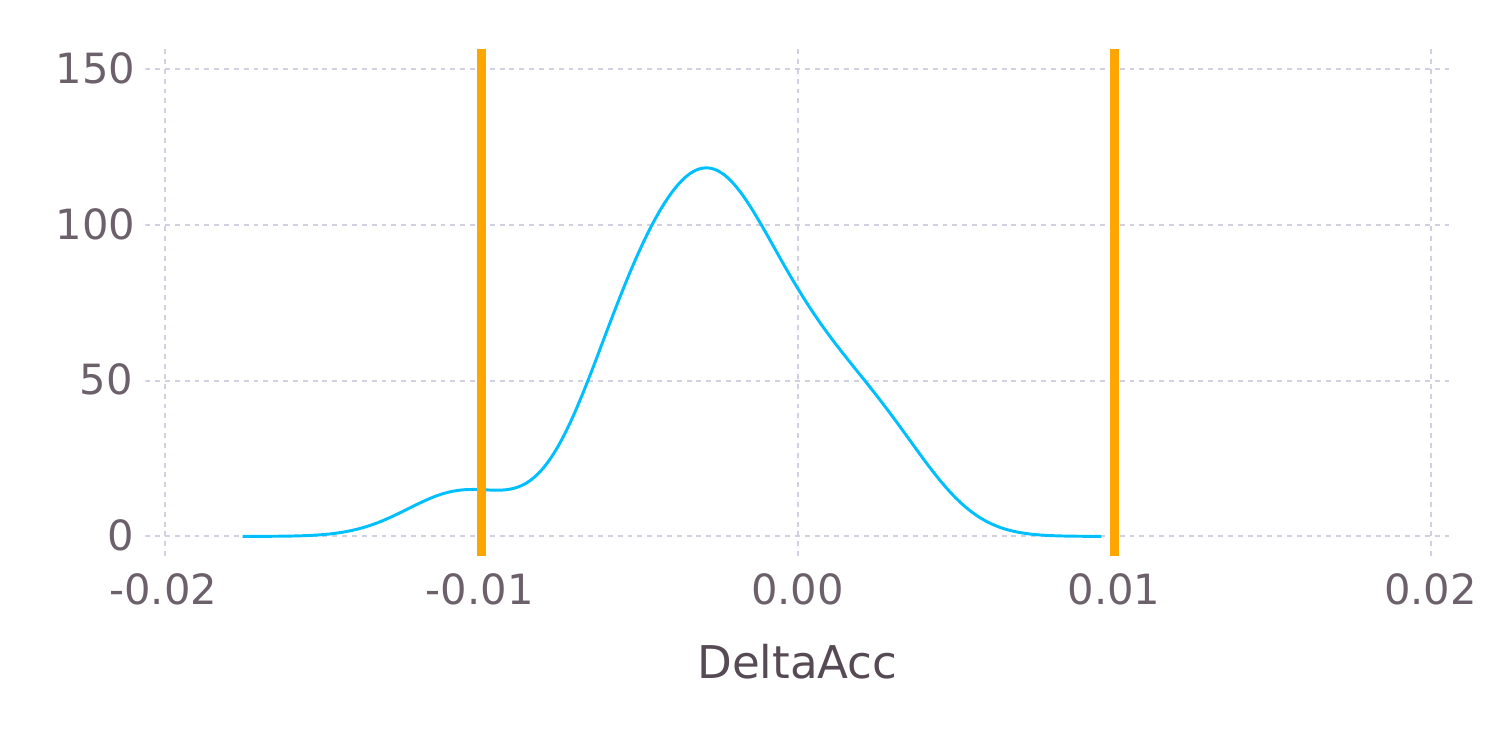} \includegraphics[width=7cm]{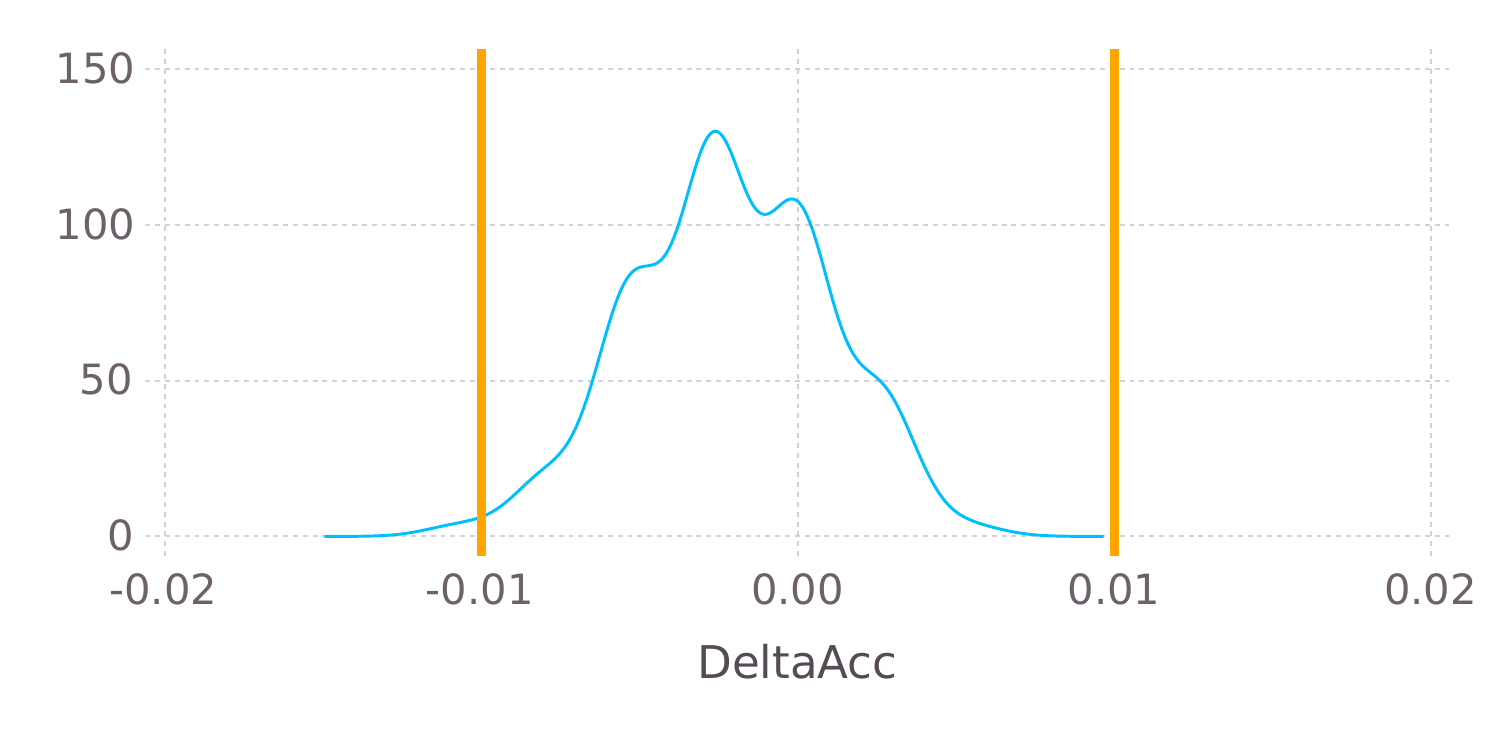}
        \caption{Density plot for the differences of accuracy between \textit{nbc} and \textit{aode} for the dataset \textit{hepatitis}
        considering only 15 of the 100 data (left) or all the data (right). Left: the null hypothesis cannot be rejected ($p=0.077>0.05$) using half the data. Right: the null hypothesis is rejected when all the data are considered
        ($p=0.048<0.05$), despite the very small effect size.}
                \label{fig:sample}
    \end{figure}

     Figure \ref{fig:sample} reports the \textit{density plots} of the differences of accuracy  between nbc and aode on the dataset \textit{hepatitis}
     in two cases: (1)  considering only $15$  of the $100$ accuracies in Table~\ref{tab:accuracies}~(left); (2) considering all the $100$ accuracies (right).  The \textit{two orange vertical lines} define the region in which the differences of accuracy is less than $1\%$---the meaning of these lines
     will be clarified in the next sections.
  
 The $p$-value is $0.077$ in the first case  and so the null hypothesis cannot be rejected.  The $p$-value becomes $0.048$ in the second case
 and so the null hypothesis can be rejected. This demonstrates how adding data leads to rejection of the null hypothesis although the difference between the two classifiers is very small in this dataset (all the mass is inside the two orange vertical lines). 
      Practical significance can be equated with the effect size, which is what the researcher is interested in. Statistical significance---the $p$-value---is not a measure of  practical significance, as shown in the example.

    \paragraph{NHST ignores magnitude and uncertainty.} {\em A very important problem with NHST is that the result of the test
    does not provide  information about the magnitude of the effect or the uncertainty of its estimate, which
    are the key information we should aim at knowing.}
    
\begin{figure}[h]
    \centering
        \includegraphics[width=7cm]{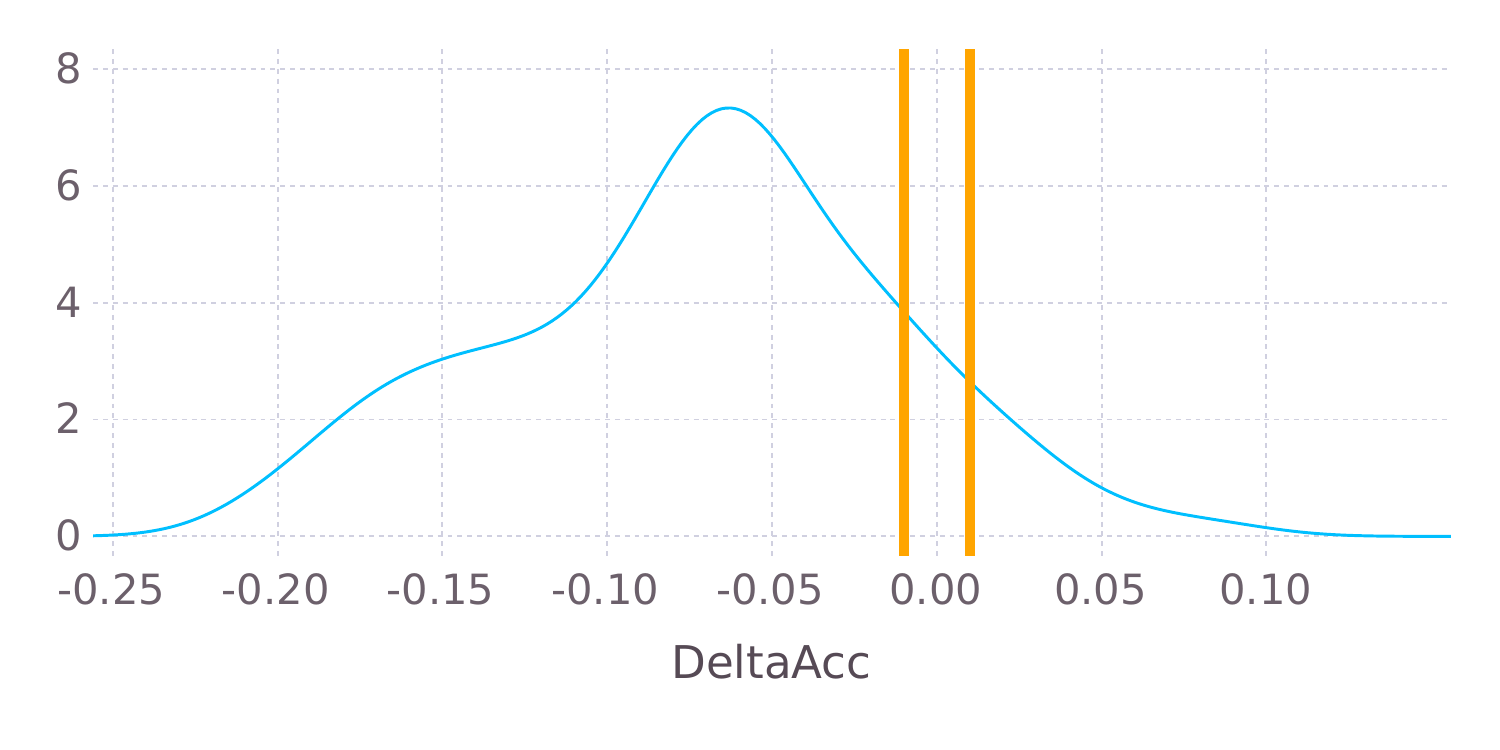}
    \includegraphics[width=7cm]{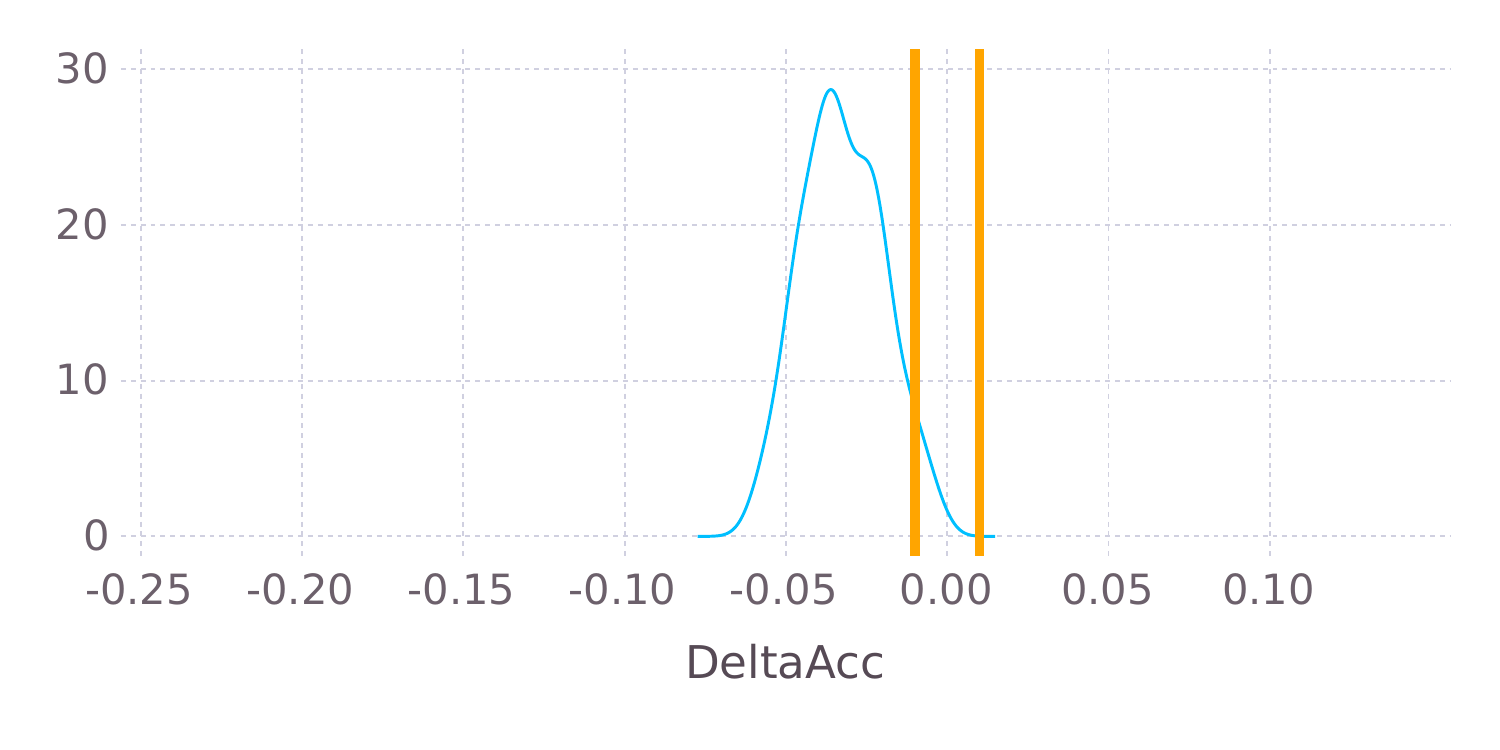}
        \caption{Density plot for the differences of accuracy (DeltaAcc) between \textit{nbc} and \textit{aode} for the datasets \textit{ecoli} (left) and \textit{iris} (right). The null hypothesis is rejected ($p<0.05$) with similar $p$-values, even though the two cases have very different uncertainty. For \textit{ecoli}, the uncertainty is very large and includes zero.}
                \label{fig:uncertainty}
    \end{figure}

A consequence of this limit is that: (i) a null hypothesis can be rejected despite a very small effect;
    (ii) a null hypothesis can be rejected even though there is a large uncertainty in the effect's magnitude and the region of uncertainty
    includes (or is  extremely close to) zero, that is, no effect.
    Figure \ref{fig:sample} (right) shows a case for which $p=0.048<0.05$ and the result is therefore declared to be statistically significant.
    However, from the density plot, it is clear that the magnitude of the effect is very small (all inside the orange vertical lines bounding the less than $1\%$ difference of accuracy     region).
    Thus rejecting a null hypothesis does not provide any information about the magnitude of the effect and whether or not the effect is trivial.
    Figure \ref{fig:uncertainty} shows two cases for which $p\approx0.001<0.05$ and the result is therefore declared to be statistically significant.
    Such $p$-values are similarly low, but the two cases are extremely different. For the dataset \textit{ecoli} (left), the differences of accuracy are spread from $0.1$ to $-0.25$
    (the magnitude of the uncertainty is very large, about $35\%$), while in the second case the data are spread from $0$ to $-0.07$, a much smaller uncertainty.
    Thus, rejecting a null hypothesis does not provide us with any information about the uncertainty of the estimate.
  
%
%
    
   \paragraph{NHST yields no information about the null hypothesis.} {\em  What can we say when NHST does not reject the null hypothesis?}

        The scientific  literature contains examples of  non-significant tests interpreted as evidence of no difference between the two methods/groups being compared.
          This is wrong since NHST cannot provide evidence in favour of the null hypothesis (see also \citet[Sec 4.1.2.2]{lecoutre2014significance}  for further examples on this point). 
  When NHST does not reject the null hypothesis, no conclusion can be made. 
  
\begin{figure}[h]
    \centering
        \includegraphics[width=7cm]{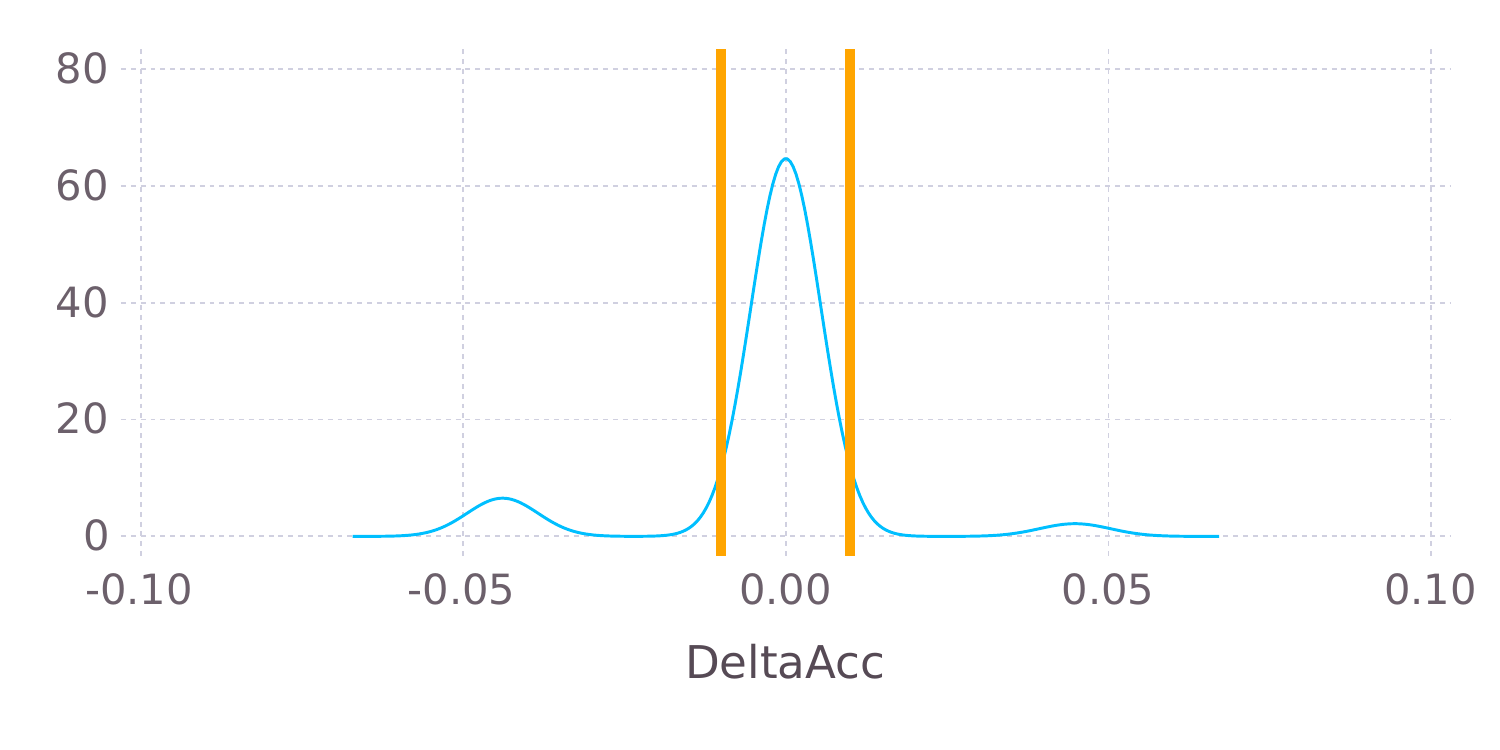}         \includegraphics[width=7cm]{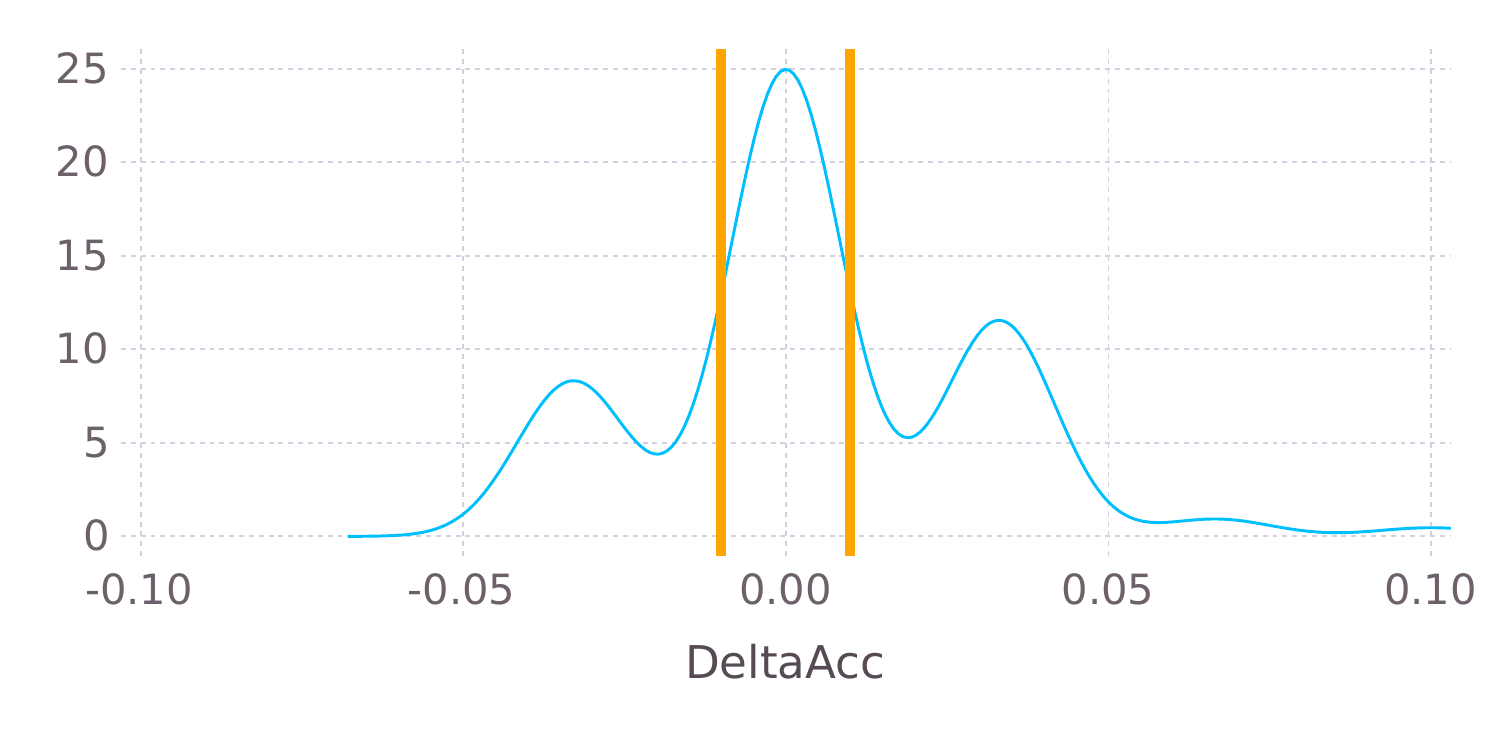}
        \caption{Density plot for the differences of accuracy between \textit{nbc} and \textit{aode} for the datasets \textit{audiology} (left)
        and \textit{breast-cancer} (right). In \textit{audiology}, we can only say that the null hypothesis cannot be rejected $p=0.622$ ($>0.05$). NHST does not allow the conclusion that the null hypothesis is true although practically
        all the data lies in the interval $[-0.01,0.01]$ (the differences of accuracy are less than $1\%$). A similar $p$-value
        corresponds to a very different situation for \textit{breast-cancer}.}
                    \label{fig:null}
    \end{figure}

  Researchers may be interested in the probability that two classifiers are equivalent. For instance, a recent paper contains the following conclusion: ``{\em there is no significant difference between (...) under the significance level of $0.05$. This is quite a remarkable conclusion.''} This is not correct, we cannot conclude anything in this case! 
    Consider for example Figure~\ref{fig:null}~(left), which shows the density plot of the differences of accuracy between nbc and aode for the dataset \textit{audiology}. The correlated t-test 
      gives a $p$-value $p=0.622$ and so it fails to reject the null hypothesis. However, NHST does not allow us to reach any conclusion about the possibility that the two classifiers may actually be equivalent in this dataset, although the majority of data  (density plot) seems to support this hypothesis.
       We tend to interpret these cases as acceptance of the null hypothesis or to even (mis)understand the $p$ value as a ``62.2\% probability that the performance of the classifiers is the same''. This is also evident from Figure \ref{fig:null} (right), where a  very similar $p$-value, $p=0.598$, corresponds to a 
       different density plot---with more uncertainty and so less evidence in favour of the null.
  
  \paragraph{There is no principled way to decide the $\alpha$ level.} In the above examples, we rejected the null hypothesis at $\alpha=0.05$. We understand this difference as significant, but not as significant as if we could reject it at $\alpha=0.01$. What is the actual meaning of this? Can we reject it at $\alpha=0.03$?
  The $\alpha$ level represents the crucial threshold to declare the experiment successful. With its importance, it needs to be set with care. However, since it is used to compare the meaningless $p$-values, $\alpha$ is equally meaningless. By using the NHST, the researcher is forced to select an important threshold with no practical meaning. Using the customary thresholds of $0.05$ and $0.01$ merely allows her/him to shift the responsibility to the unsubstantiated traditional habit.

  We pretend that $\alpha$ is the proportion of cases in which $H_0$ would be falsely rejected if the same experiment was repeated. This would be true if the $p$-value represented the probability of the null hypothesis. In reality, $\alpha$ is the proportion of cases in which experiments yield the data that is more extreme than expected under $H_0$; the actual probability of falsely rejecting $H_0$ is also related to the probability of $H_0$ and the data.

  For $\alpha$ to be meaningful, it would need to set the required effect size or at least the probability of the hypothesis, not the likelihood of the data.

  \paragraph{The inference depends on the sampling intention.}
  Consider analysing a data set of $n$ observations with a NHST test.
  The sampling distribution used to determine the critical value of the test 
  assumes that our intention was to collect exactly $n$ observations.
  If the intention was different---for instance in machine learning you typically compare two algorithms on all the datasets that are available---, the sampling distribution changes to reflect the actual sampling intentions \citep{kruschke2010bayesian}.
  This is never done, given the difficulty of formalizing one's intention and of devising an 
  appropriate sampling distribution.
  This problem is thus important but generally ignored. 
  Thus for the data set the hypothesis test (and thus the p-value)
  should be computed differently, depending on the intention of the person who collected the data
  \citep{kruschke2010bayesian}.

  \section{Bayesian analysis of experimental results}
  \label{sec:student}
  There are two main Bayesian approaches for the analysis of  experimental results.
The first Bayesian approach, as NHST, is also based on a null value. 
The analyst has to set up two competing models of what values are possible. 
One model assumes that only the null value is possible.
The alternative model assumes a broad range of other values is also possible. 
Bayesian inference is used to compute which model is more credible, given the data. 
This method is called \textit{Bayesian model comparison} and uses so-called ``Bayes factors'' \citep{berger1985,aitkin1991posterior,kass1995bayes,berger1996intrinsic}.

The second Bayesian approach does not set any null value.
The analyst simply has  to set up a range of candidate values (prior model), including the zero effect, and use Bayesian inference to compute the relative credibilities of all the candidate values (the posterior distribution). This method is called \textit{Bayesian estimation} \citep{gelman2014bayesian,kruschke2015doing}.

The choice of the method depends on the specific question that the analyst aims at answering, but in machine learning the estimation approach is usually preferable because it provides richer information to the analyst.
For this reason, we will focus on the Bayesian  estimation approach that hereafter we will simply call \textit{Bayesian analysis}.

The first step in Bayesian analysis is establishing a
descriptive mathematical model of the data. 
In a parametric model, this mathematical model 
is the the \textit{likelihood function} that provides the
probability of the observed data for each candidate value of
the parameter(s) $p(Data|\theta)$.
The second step is to establish the credibility for each value of the parameter(s) before observing 
data, the \textit{prior distribution} $p(\theta)$.
The third step is to use Bayes' rule to combine likelihood and prior to obtain the 
\textit{posterior distribution} of the parameter(s) given the data $p(\theta|Data)$.
The questions we pose in  statistical analysis can be answered by querying this posterior distribution in different ways.

As a concrete example of Bayesian analysis we will compare the accuracies of two competing classifiers via cross-validation on multiple data sets (Table \ref{tab:accuracies}). For this purpose, we will adopt the  \textit{correlated Bayesian $t$-test} proposed by \cite{coraniML2015}.

  \begin{SnugshadeB}
\begin{MethodB}[correlated t-test]
   The Bayesian correlated t-test  is used for the analysis of cross-validation results on a single dataset and it accounts for the correlation due to the overlapping training sets.
   The test is based on the following (generative) model of the data:
  \begin{equation}
  \mathbf{x}_{\scriptscriptstyle n\times 1}=\mathbf{1}_{\scriptscriptstyle n\times 1} \mu+\mathbf{v}_{\scriptscriptstyle n\times 1}, \label{eq:regression-model} 
  \end{equation}
  where $\bm{x}=(x_1,x_2,\dots,x_n)$  is the vector of differences of accuracy, $\mathbf{1}_{\scriptscriptstyle n\times 1}$ is a vector of ones, $\mu$ is the parameter of interest
  (the mean difference of accuracy) and  $\mathbf{v}\sim \mathrm{MVN}(0,\bm{\Sigma}_{\scriptscriptstyle n\times n})$ is a multivariate Normal noise with zero mean
  and covariance matrix $\bm{\Sigma}_{\scriptscriptstyle n\times n}$.
   The covariance matrix $\bm{\Sigma}$  is characterized as follows: $\bm{\Sigma}_{ii}=\sigma^2$ and $\bm{\Sigma}_{ij}=\sigma^2 \rho$ for all $i\neq j\in 1,\dots,n$,
   where $\rho$ is the correlation and $\sigma^2$ is the variance and, therefore, the covariance matrix takes into account the correlation due to cross-validation.  
  Hence, the likelihood model of data is 
    \begin{equation}
  \begin{array}{l}
  p(\mathbf{x}|\mu,\bm{\Sigma})=\dfrac{\exp(-\frac{1}{2}(\mathbf{x}-\mathbf{1}\mu)^{T}\bm{\Sigma}^{-1}(\mathbf{x}-\mathbf{1}\mu))}{(2\pi)^{n/2}\sqrt{|\bm{\Sigma}|}}.\vspace{2mm}\\
  \end{array}\label{eq:lik-correlated}
  \end{equation}
  The likelihood (\ref{eq:lik-correlated}) does not allow to estimate $\rho$ from data, since the maximum likelihood estimate
  of $\rho$ is $\hat{\rho}=0$ regardless the observations \citep{coraniML2015}. 
  This confirms that $\rho$ is not identifiable: thus the Bayesian correlated t-test
  adopts the same heuristic $\rho=\frac{n_{te}}{n_{tot}}$ suggested by \cite{nadeau2003inference}. 
  
 In Bayesian estimation, we aim at estimating the unknown parameters $\mu,\nu=1/\sigma^2$ and in particular $\mu$, which is the parameter of interest
  in the Bayesian correlated t-test.
  To this end, we consider the following prior:
  \begin{align*}
  p(\mu,\nu|\mu_0,k_0,a,b)=N\left(\mu;\mu_0,\frac{k_0}{\nu}\right)G\left(\nu;a,b\right)=NG(\mu,\nu;\mu_0,k_0,a,b),
  \label{eq:normal-gamma}
  \end{align*}
  which is a Normal-Gamma distribution \cite[Chap.~5]{bernardo2009bayesian} with parameters $(\mu_0,k_0,a,b)$.
  The Normal-Gamma prior is conjugate  to the likelihood  (\ref{eq:lik-correlated}). If we choose the prior parameters $\{\mu_0=0$, $k_0 \rightarrow \infty$, $a=-1/2$, $b=0\}$ (matching prior), the resulting posterior distribution of $\mu$ is the following Student distribution:
  \begin{equation}
  p(\mu|\bm{x},\mu_0,k_0,a,b)=St\left(\mu;n-1,\bar{x},\left(\frac{1}{n}+\frac{\rho}{1-\rho}\right)\hat{\sigma}^2\right),
  \label{eq:student-bayesian}
  \end{equation}
  where $\bar{x}=\tfrac{\sum_{i=1}^n x_i}{n}$ and $\hat{\sigma}^2=\tfrac{\sum_{i=1}^n (x_i-\bar{x})^2}{n-1}$.
  For these values of the prior parameters, the posterior distribution of $\mu$ (\ref{eq:student-bayesian}) coincides 
  with the Student distribution used in the frequentist correlated t-test in (\ref{eq:freqstud}).
  For instance, consider  the first data set in Table \ref{tab:accuracies}, we have that $\bar{x}=-0.0194$, $\hat{\sigma}=0.01583$, $\rho=1/10$, $n=100$ and so
   $p(\mu|\bm{x},\mu_0,k_0,a,b)=St\left(\mu;99,-0.0194,0.000030\right)$.
  The output of the Bayesian analysis is the posterior of $\mu$, $p(\mu|\bm{x},\mu_0,k_0,a,b)$, which we can plot and query.
   
%
 \end{MethodB}
 \end{SnugshadeB}
 In (\ref{eq:student-bayesian}), we have reported the posterior distribution obtained under the matching prior---for which  the  probability of the Bayesian correlated $t$-test and the $p$-value of the frequentist correlated $t$-test are numerically equivalent. Our aim is to show that although they are numerically equivalent the inferences drawn by the two approaches  are very different. In particular we will show  that a different interpretation of the same numerical value can completely change our prospective and allow us to make informative decisions. In other words, in this case  \textbf{the cassock does make the priest}!
  

  \subsection{Comparing nbc and aode through Bayesian analysis: a colour thinking}
      
\begin{figure}[h]
  \begin{center}
            \includegraphics[width=8cm]{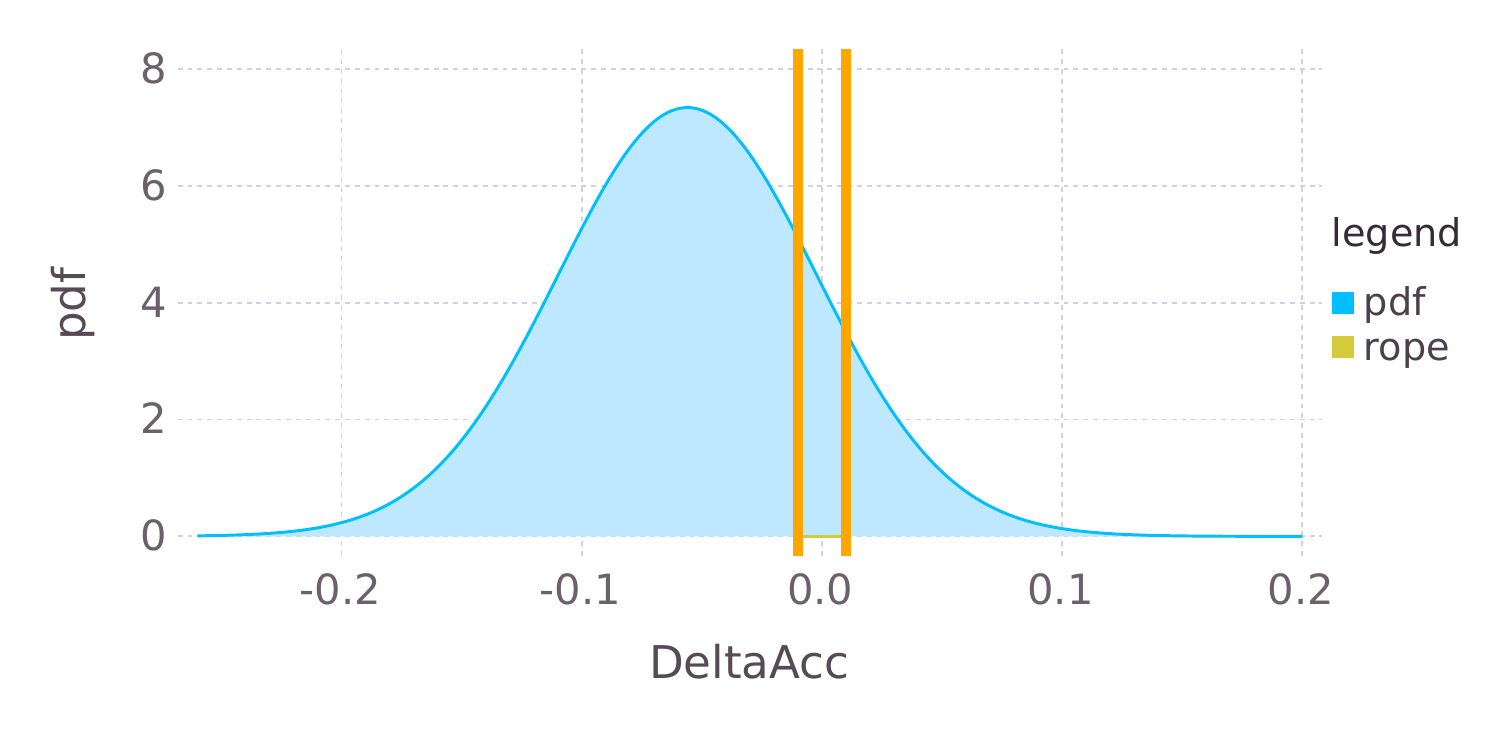}\\
  \end{center}
  \caption{Posterior of the Bayesian correlated t-test for the difference between \textit{nbc} and \textit{aode} in the dataset \textit{squash-unsorted}. }
      \label{fig:postaud}
\end{figure}

  Consider the dataset \textit{squash-unsorted}, with the posterior computed by the Bayesian correlated t-test for the difference between \textit{nbc} and \textit{aode}, 
 as shown in Figure \ref{fig:postaud}. The vertical orange lines mark again the region corresponding to a difference of accuracy  of less than $1\%$ (we
  will clarify the meaning of this region later in the section).
  In Bayesian analysis, the experiment is summarized by the posterior distribution (in this case a Student distribution).
  The posterior describes the distribution of the mean difference of accuracies between the two classifiers.

  By querying the posterior distribution, we can evaluate the probability of the hypothesis.
  We can for instance infer the probability that \textit{nbc} is better/worse than \textit{aode}.
  Formally $P(nbc>aode)=0.165$ is the integral of the posterior distribution from zero to infinity or, equivalently, the posterior probability that the mean of the differences of accuracy   between \textit{nbc} and \textit{aode} is greater than zero.  $P(aode>nbc)=1-P(nbc>aode)=0.835$ is the integral of the posterior between minus infinity and zero or, equivalently, the posterior probability that the mean of the differences of accuracy is less than zero. 
  
  
  Can we say anything about the probability that \textit{nbc} is practically equivalent to \textit{aode}? Bayesian analysis can answer this question.
  First, we need to define the meaning of ``practically equivalent''. In classification, it is sensible to define that two classifiers whose mean difference 
  of accuracies is less that $1\%$ are practically equivalent. The interval $[-0.01,0.01]$ thus defines a \textit{region of practical equivalence} (rope) \citep{kruschke2015bayesian}
  for classifiers.\footnote{In classification $1\%$ seems to be a reasonable choice. However, in other domains a different value could be more suitable.} Once we have defined a rope,  from the posterior we can compute the probabilities:
  \begin{itemize}
	  \item $P(nbc\ll aode)$: the posterior probability of the mean difference 
  of accuracies being   \textit{practically} negative, namely the integral of the posterior on the interval $(-\infty,-0.01)$.
	  \item $P(nbc=aode)$: the posterior probability of the two classifiers being practically equivalent, namely the integral of the posterior over the rope interval.
	  \item $P(nbc\gg aode)$: the posterior probability of the mean difference 
  of accuracies being  	  \textit{practically} positive, namely the integral of the posterior on the interval 
	  ($0.01,\infty$).
  \end{itemize}
  $P(nbc=aode)=0.086$ is the integral of the posterior distribution between the vertical lines (the rope) shown in Figure \ref{fig:postaud} 
  and it represents the probability that the two classifiers are practically equivalent. Similarly, we can compute
  the probabilities that the two classifiers are practically different, which are $P(nbc \ll aode)=0.788$
  and  $P(nbc \gg aode)=0.126$.

\begin{figure}[h]
  \begin{center}
            \includegraphics[height=3.6cm]{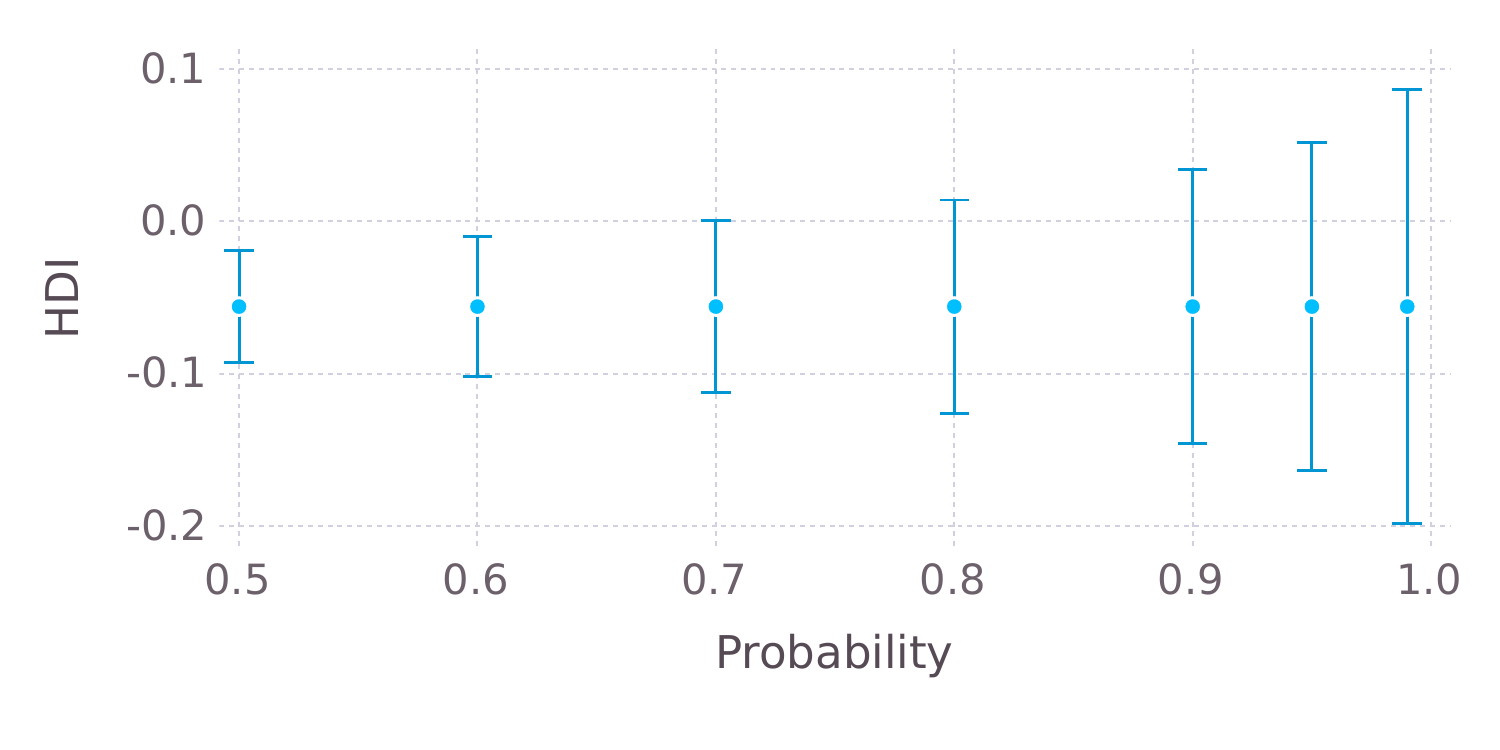} 
  \end{center}
  \caption{Posterior HDIs of the Bayesian correlated t-test for the difference between \textit{nbc} and \textit{aode} in the dataset \textit{squash-unsorted}. }
      \label{fig:postaudhdi}
\end{figure}

The posterior also shows the uncertainty in the estimate, because the distribution shows the relative
credibility of values across the continuum. One way to summarize the uncertainty is by marking the span of values that are the most
credible and cover $q\%$ of the distribution (e.g., $q=90\%$).
These are called  the \textit{High Density Intervals} (HDIs) and they are shown in Figure \ref{fig:postaudhdi} (center)
for $q=50,60,70,80,90,95,99$. 

\begin{figure}[h]
    \centering
            \includegraphics[width=7cm]{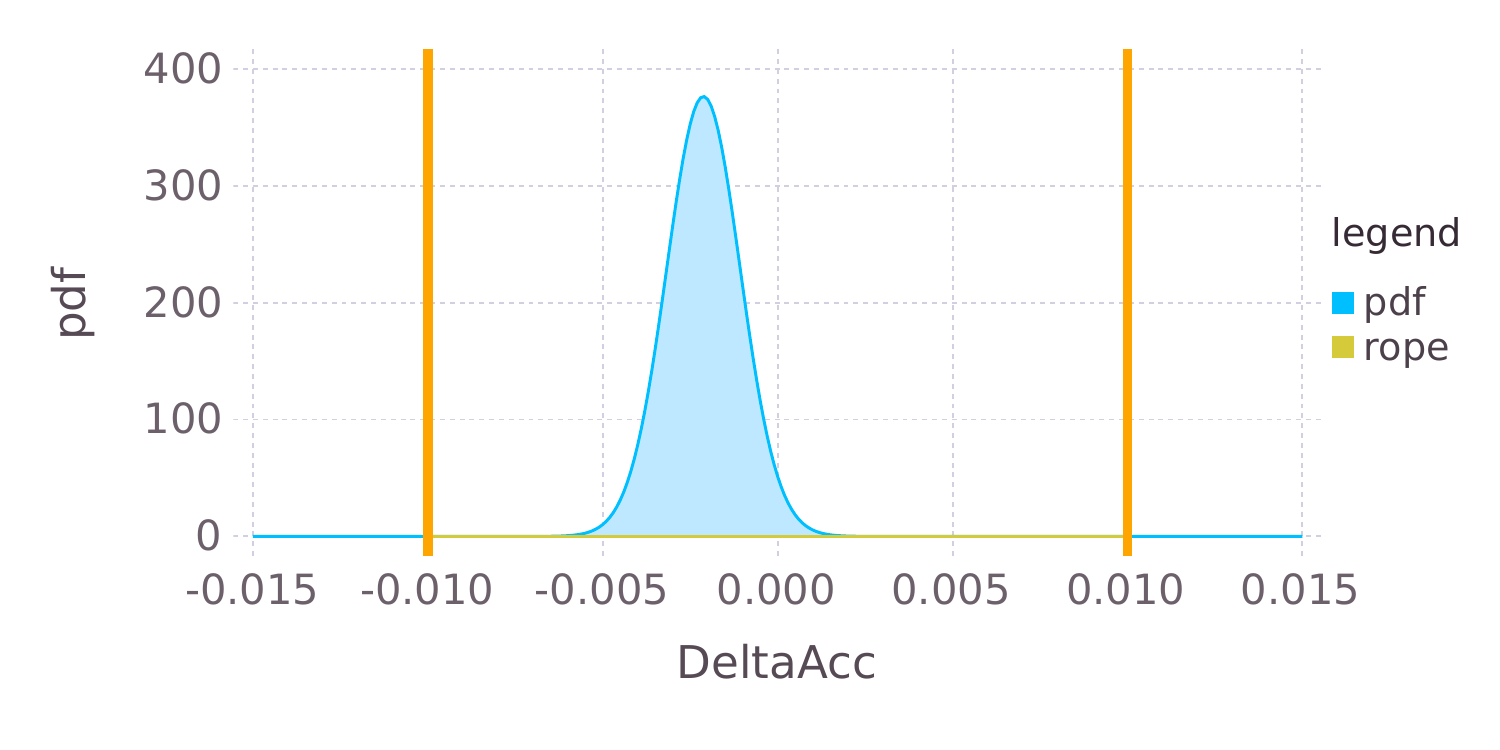}\\
    \includegraphics[width=7cm]{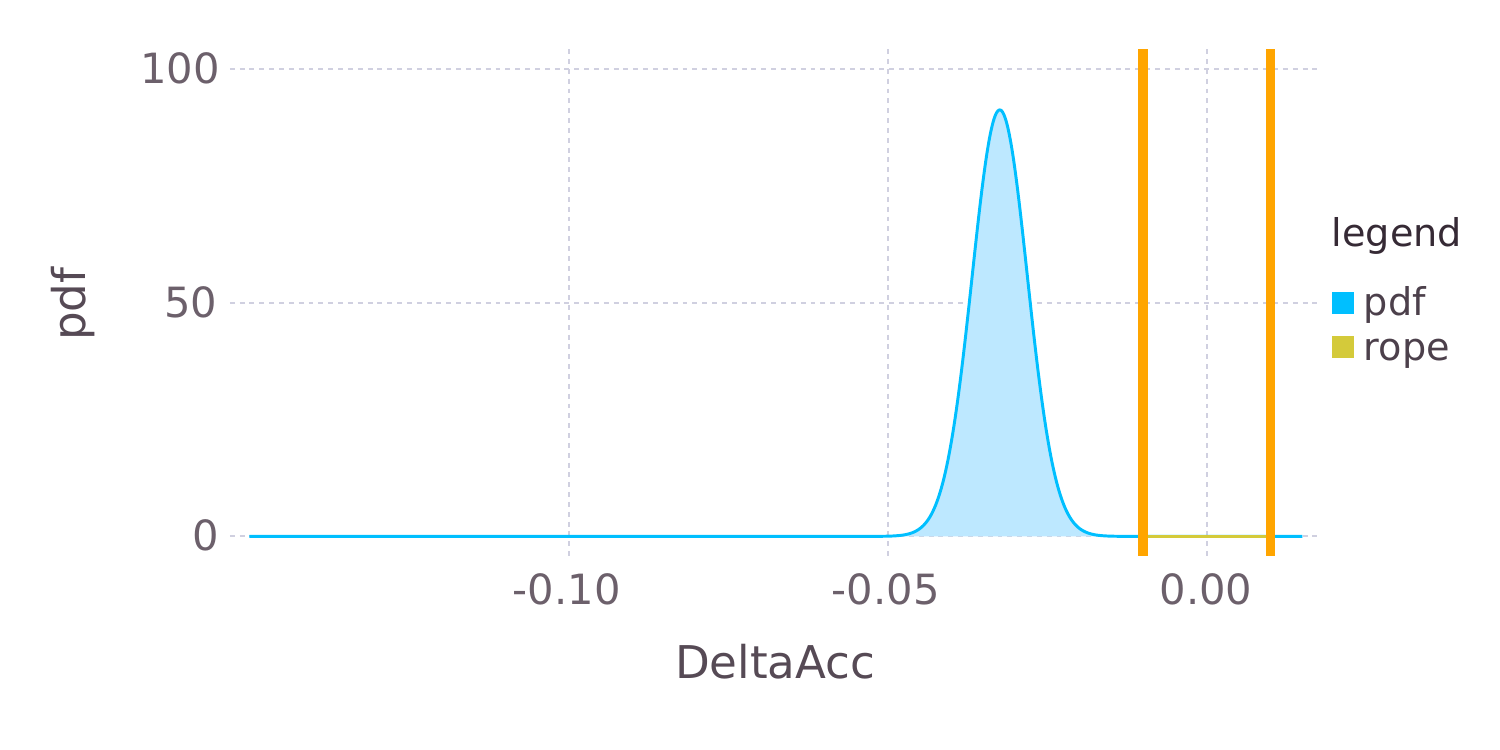}
            \includegraphics[width=7cm]{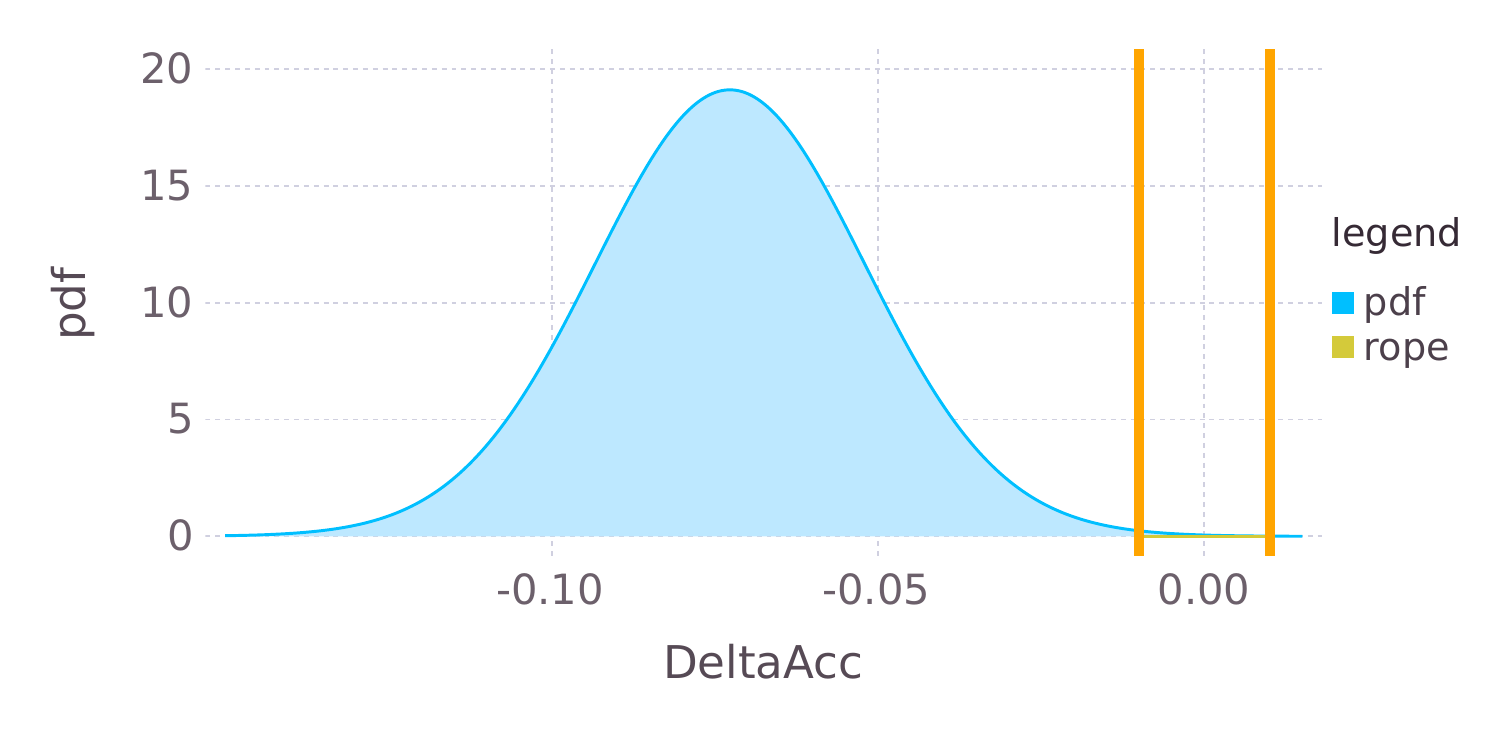}
        \caption{Posterior  for  \textit{nbc} vs.\ \textit{aode} in \textit{hepatitis} (top), \textit{ecoli} (left) and \textit{iris} (right).}
                \label{fig:BAmagnuncert}
    \end{figure}

Thus the posterior distribution equipped with rope:  
\begin{enumerate}
 \item estimates the posterior probability of a sensible null hypothesis (the area within the rope);
  \item claims significant differences that also have a practical meaning (the area outside the rope);
 \item represents magnitude (effect  size) and uncertainty (HDIs);
 \item does not depend on the sampling intentions.
\end{enumerate}
To see that, we apply Bayesian analysis to the critical cases for  NHST presented in the previous section.
Figure \ref{fig:BAmagnuncert} shows the posterior for \textit{hepatitis} (top), \textit{ecoli} (left) and \textit{iris} (right).
For \textit{hepatitis}, all the probability mass is inside the \textit{rope} and so we can conclude that 
 \textit{nbc} and \textit{aode} are practically equivalent: $P(nbc=aode)=1$.
 For \textit{ecoli}  and \textit{iris}, the probability mass is all in $(-\infty,0]$ and so $nbc<aode$.
 However, the posterior gives us more information:  there is much more uncertainty in the \textit{iris} dataset.
 The posteriors thus provide us the same information as the density plots shown in Figures~\ref{fig:sample}--\ref{fig:null}.

\begin{figure}[h]
    \centering
        \includegraphics[width=7cm]{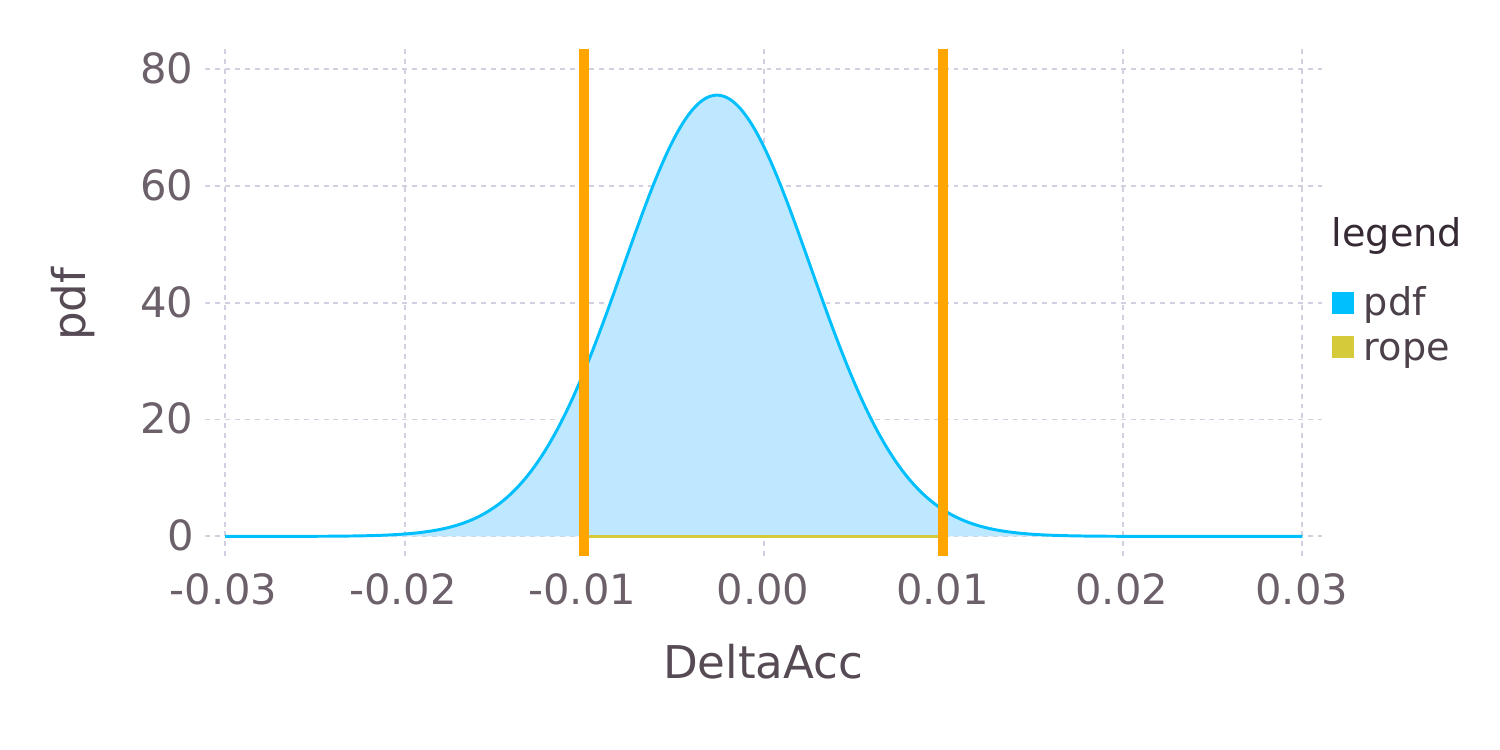}         \includegraphics[width=7cm]{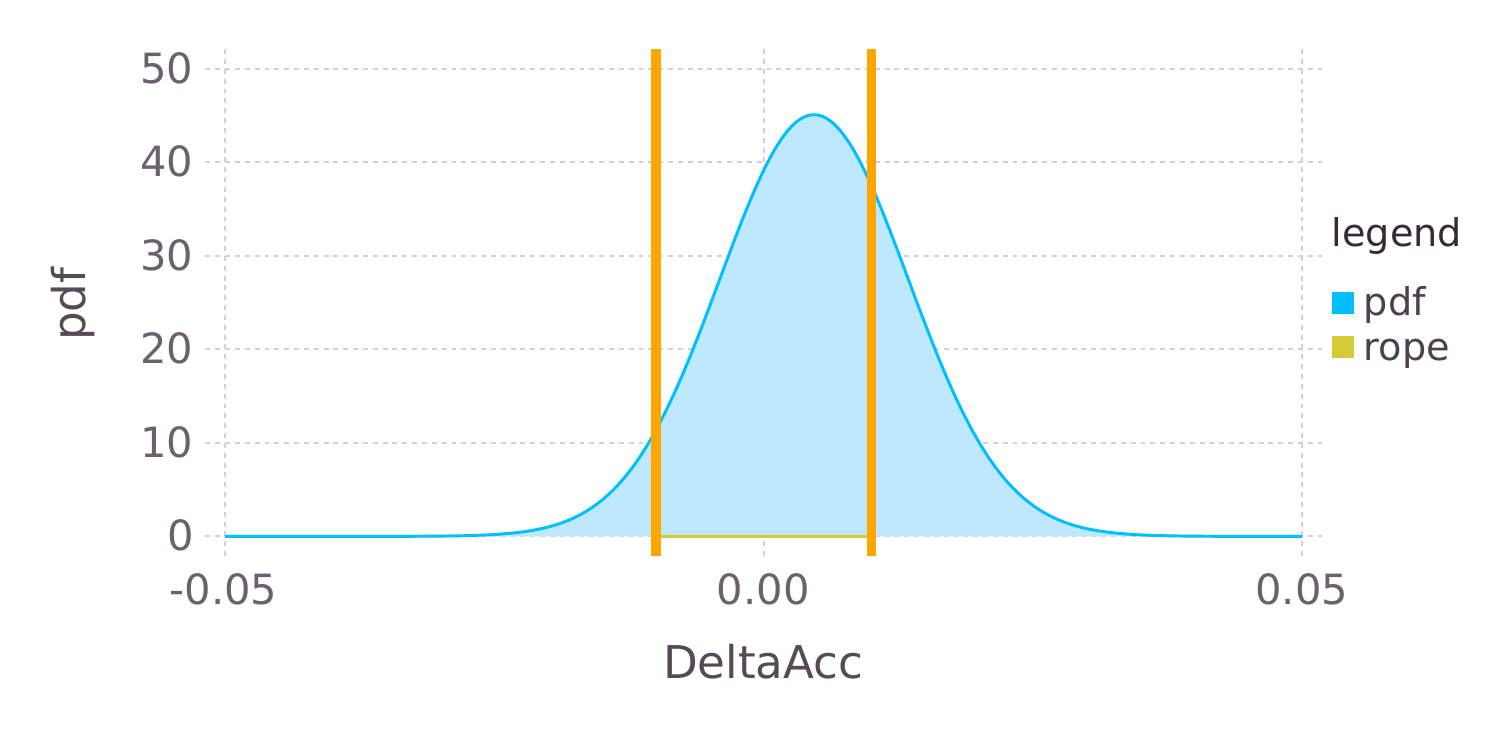}
        \caption{Posterior for  \textit{nbc} vs.\ \textit{aode} in \textit{audiology} (left)
        and \textit{breast-cancer} (right).}
                    \label{fig:BAnull}
    \end{figure}

Consider now Figure \ref{fig:BAnull}; those are two examples for which we cannot clearly say whether \textit{nbc} and \textit{aode} 
are practically equivalent or practically different. We cannot decide in an obvious way and this is evident from the posterior.
Compare these figures with Figure \ref{fig:postaud},  which is similar. 
  
\begin{figure}[h]
    \centering
        \includegraphics[width=15cm]{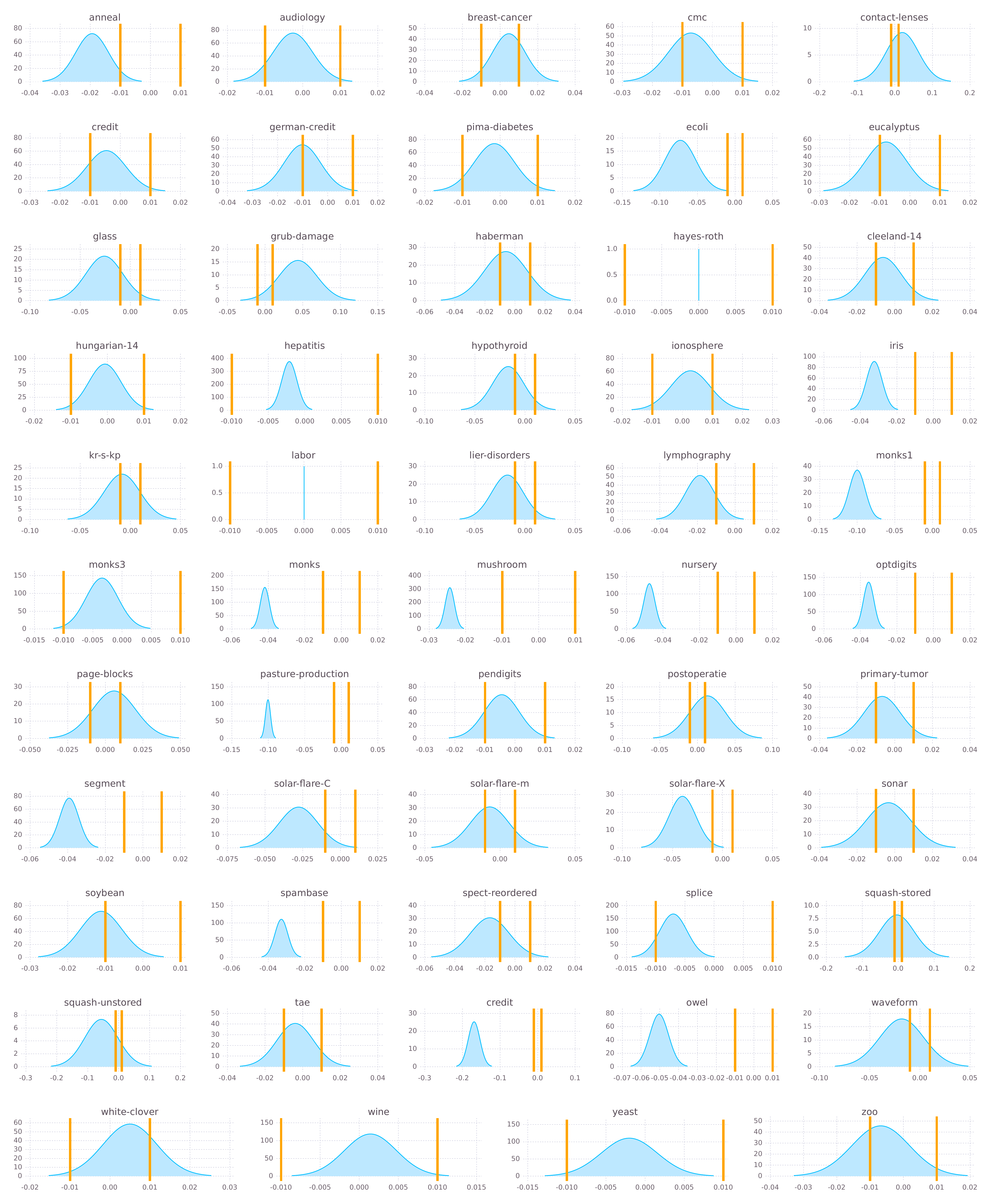}        
        \caption{Posteriors of \textit{nbc} versus \textit{aode} for all $54$ datasets.}
                    \label{fig:piechartbayall}
    \end{figure}

Let us repeat the previous  analysis for all  $54$ datasets: Figure~\ref{fig:piechartbayall} reports the 
posteriors for \textit{nbc} versus \textit{aode} in all those cases.
Looking at the posteriors, we see that there are $12$ cases where
\textit{aode} is  practically better than \textit{nbc}, since all the posterior is outside (to the left of) the rope (in the datasets \textit{ecoli, iris, monks1, monks, mushroom, nursey, optdigits, pasture, segment, spambase, credit, owel}). There are  $6$ datasets where  \textit{nbc} and \textit{aode} are practically equivalent (\textit{hayes-roth}, \textit{hungarian14}, \textit{hepatitis}, \textit{labor}, \textit{wine}, \textit{yeast}), since  the entire posterior is inside the rope.
We see also that there are no cases where \textit{nbc} is practically better than \textit{aode} (posterior to the right of the rope).
The posteriors give us information about the magnitude of effect size, practical difference and equivalence, as well as the related uncertainty.

\subsection{Sensible automatic decisions}
In machine learning, we often need to perform many analyses. So it may be convenient to devise a tool for automatic decision from the posterior.
This means that we have to summarize the posterior in some way, with a few numbers. 
However, we must be aware that every time we do that we go back to that sort of black and white analysis
whose limitations have been discussed before. In fact, by summarizing the posterior, we lose information,
but we can do so in a conscious way.
We have already explained the advantages of introducing a \textit{rope}, so we can make automatic
decisions based on the three probabilities $P(nbc\ll aode)$, $P(nbc=aode)$ and $P(nbc\gg aode)$. 
In this way, we lose information but we introduce shades in the black and white thinking.
$P(nbc\ll aode)$, $P(nbc=aode)$ and $P(nbc\gg aode)$ are probabilities and their interpretation is clear.
$P(nbc=aode)$ is the area of the posterior within the rope and represents  the probability that \textit{nbc} and \textit{aode}
are practically equivalent. $P(nbc\ll aode)$ is the area of the posterior to the left of the rope 
and corresponds to the probability that \textit{nbc} is practically better than \textit{aode}.
Finally, $P(nbc\gg aode)$ is the area to the right of the rope and represents the probability that \textit{aode} is practically better than \textit{nbc}.
Since these are  the actual probabilities of the decisions we are interested in, in classification, we need not think in terms of Type-I errors to make decisions.
We can simply  make decisions using these  probabilities, which have a direct interpretation---contrarily to $p$-values.  For instance we can decide 
\begin{enumerate}
 \item $nbc\ll aode$ if $P(nbc\ll aode)>0.95$;
\item  $nbc\gg aode$ if $P(nbc\gg aode)>0.95$;
\item  $nbc=aode$ if $P(nbc=aode)>0.95$.
\end{enumerate}
We can also decide with a probability of, for instance, $0.90$ or $0.80$ (if this is appropriate in the given context).  Table~\ref{tab:nbcaodetab} compares the Bayesian decisions based on the above rule with the NHST decisions.   
The NHST fails to reject the null in $35/54$ datasets; Bayesian analysis declares
the two classifiers equivalent in $6$ of such datasets. Conversely, when NHST rejects the null (in $19/54$ datasets), the Bayesian analysis declares
nbc$\ll$aode or nbc$\gg$aode in $14$ cases, nbc$=$aode in $1$ case and no decision in $4$ cases.
Overall, Bayesian analysis allows us to make a decision in $6+1+14=21$ datasets, while NHST makes a decision only in $19$ cases.

             \begin{table}[h]
      \rowcolors{1}{white}{lightblue}
  \centering
  \begin{tabular}{@{}lccc@{}}
  & \multicolumn{3}{c}{\textit{When NHST does not reject the null}}  \\ \toprule
  pair & Data sets & \multicolumn{2}{c}{Bayesian decision}  \\ 
    & (out of 54) & nbc$=$aode & No decision  \\ \midrule
  nbc-aode & 35 & 6 & 29  \\
  \end{tabular}\\
  \vspace{6mm}
        \rowcolors{1}{white}{lightblue}
  \begin{tabular}{@{}lcccc@{}}
  & \multicolumn{4}{c}{\textit{When NHST rejects the null}} \\ \toprule
  pair & \multicolumn{1}{l}{\begin{tabular}[c]{@{}c@{}}Data sets\\ 
  (out of 54)\end{tabular}} & \multicolumn{3}{c}{Bayesian decision}  \\ 
  pair & \multicolumn{1}{l}{\begin{tabular}[c]{@{}c@{}}\end{tabular}} & \multicolumn{1}{c}{nbc$=$aode} & \multicolumn{1}{c}{nbc$\ll$aode or nbc$\gg$aode} & \multicolumn{1}{c}{No decision} \\ \midrule
  nbc-aode 	& 19 & 1 & 14 & 4 \\
  \end{tabular}
  \caption{Results referring to comparisons in which the NHST \textit{rejects} the null.}
    \label{tab:nbcaodetab}
  \end{table}

  Sometimes also in Bayesian analysis it may be convenient to think in terms of errors. We can easily do that by defining a loss function.
  A loss function defines the loss we incur in making the wrong decision. 
  The decisions we can take are nbc$\ll$aode (denoted as $a_l$), nbc$\gg$aode  ($a_r$), nbc$=$aode ($a_c$) or none of them ($a_n$). Consider for instance the following loss matrix. 
  \begin{equation}
\begin{matrix}
 a_l\\
 a_c\\
 a_r\\
 a_n
\end{matrix}
\stackrel{a_l~~~~~a_c~~~~~a_r}{
\left[\begin{matrix}
 0 & 20 & 20 \\
 20 & 0 & 20 \\
 20 & 20 & 0 \\
 1 & 1 & 1 \\
\end{matrix}\right]}
  \end{equation} 
The first row gives us the loss we incur in deciding $a_l$
when $a_l$ is true (zero loss),  $a_c$ is true (loss is $20$) and $a_r$ is true (loss is $20$).
Similarly for the second and third rows.
The last row is the loss we incur in making no decision.
The expected loss can be simply obtained by multiplying the above matrix $L$ for the vector of posterior
probabilities of nbc$\ll$aode, nbc$=$aode and nbc$\gg$aode ($p=[p_l,p_c,p_r]^T$), i.e., 
$Lp$. The row  of $Lp$ corresponding to the lowest loss determines the final decision.
Since $0.05\cdot20=1$, this leads to the same decision rule discussed previously ($P(\cdot)> 0.95$).

  \subsection{Comparing NHST and Bayesian analysis for other classifiers}
  In this section we extend the previous analysis to other classifiers besides nbc and aode: hidden naive
  Bayes (hnb), j48 decision tree (j48) and j48 grafted (j48-gr).

%

  The aim of this section is to show that the pattern described above is general 
  and it also holds for other classifiers. The results are presented in two tables.
  First we  report on the cases in which NHST does not reject the null (Tab.~\ref{tab:rope-each-dset2}).
  Then we report on the comparisons in which the NHST rejects the null (Tab.~\ref{tab:rope-each-dset}).

  The NHST test does not reject the null hypothesis (Tab.~\ref{tab:rope-each-dset2}) in 341/540 comparisons.
  In these cases the NHST does not make any conclusion: it cannot tell
  whether the null hypothesis is true\footnote{A point null is however always false, as already discussed.} or whether is false but the evidence is too weak to reject it.

  By applying the Bayesian correlated t-test with rope and taking decisions as discussed
  in the previous section, we can draw more informative conclusions.
  In 74/341=22\% of the rejections failed by NHST, the posterior probability of the rope
  is larger than $0.95$, allowing to declare that the two analyzed classifiers are practically equivalent.
  In the remaining cases, no conclusion can be drawn with probability $0.95$.

  The rope thus provides a sensible null hypothesis which can be accepted on the basis of the data.
  When this happens we conclude that the two classifiers are  practically equivalent. This is impossible with the NHST. 

  \begin{table}[!ht]
      \rowcolors{1}{white}{lightblue}
  \centering
  \begin{tabular}{@{}lrrr@{}}
  & \multicolumn{3}{c}{\textit{When NHST does not reject the null}}  \\ \toprule
  pair & Data sets & \multicolumn{2}{c}{Bayesian decision}  \\ 
    & (out of 54) & P(rope) \textgreater .95 & No decision  \\ \midrule
  nbc--aode & 35 & 6 & 29  \\
  nbc--hnb & 30 & 0 & 30  \\
  nbc--j48 & 27 & 2 & 25 \\
  nbc--j48gr & 27 & 2 & 25  \\
  aode--hnb & 40 & 6 & 34  \\
  aode--j48 & 33 & 6 & 27  \\
  aode--j48gr & 35 & 6 & 29  \\
  hnb--j48 & 32 & 3 & 29  \\
  hnb--j48gr & 32 & 3 & 29  \\
  j48--j48gr & 50 & 40 & 10  \\ \midrule
  \textbf{total} & 341 & 74  & 267 \\ 
  \textbf{rates} & & 74/341= 0.22 & 267/341=0.78  \\ \bottomrule
  \end{tabular}
  \caption{Results referring to comparisons in which the NHST correlated t-test does not reject the null.}
\label{tab:rope-each-dset2}
  \end{table}

  \begin{table}[!ht]
      \rowcolors{1}{white}{lightblue}
  \centering
  \begin{tabular}{@{}lcccc@{}}
  & \multicolumn{4}{c}{\textit{When NHST rejects the null}} \\ 
  pair & \multicolumn{1}{c}{\begin{tabular}[c]{@{}c@{}}Data sets\\ \toprule
  (out of 54)\end{tabular}} & \multicolumn{3}{c}{Bayesian decision}  \\ 
  pair & \multicolumn{1}{c}{\begin{tabular}[c]{@{}c@{}}\end{tabular}} & \multicolumn{1}{c}{rope} & \multicolumn{1}{c}{difference} & \multicolumn{1}{c}{no decision} \\ \midrule
  nbc--aode 	& 19 & 1 & 14 & 4 \\
  nbc--hnb 	& 24 & 0 & 19 & 5\\
  nbc--j48 	& 27 & 0 & 20 & 7\\
  nbc--j48gr 	& 27 & 0 & 21 & 6 \\
  aode--hnb 	& 14 & 1 & 6 & 7\\
  aode--j48 	& 21 & 1 & 14 & 6\\
  aode--j48gr & 19 & 1 & 13 & 5\\
  hnb--j48 	& 22 & 0 & 17 & 5\\
  hnb--j48gr 	& 22 & 0 & 17 & 5\\
  j48--j48gr 	& 4 & 2 & 1 & 1\\ \midrule
  \textbf{total} & 199 & 6 & 142 & 51 \\ 
  \textbf{rates} & & 6/199= 0.03 & 142/199=0.71 & 51/199=0.26\\ \bottomrule
  \end{tabular}
  \caption{Results referring to comparisons in which the NHST \textit{rejects} the null.}
  \label{tab:rope-each-dset}
  \end{table}


  Let us consider now the comparisons in which NHST claims significance (Tab.~\ref{tab:rope-each-dset}). 
  There are 199 such cases.
  The Bayesian estimation procedure
  confirms the significance of 142/199 (71\%) of them: in these cases
  it estimates either $P(left)>0.95$ or $P(right)>0.95$.\footnote{In the comparison nbc-aode,  left means $nbc\ll aode$ and right $nbc\gg aode$.}
  In these cases the accuracy of the two compared classifiers are \textit{practically} different.
  In 51/199 cases (26\%) the Bayesian test does not make any conclusion. This means that a sizeable amount of probability lies within the rope, despite the statistical significance claimed by the frequentist test. 
  In the remaining cases (6/199=3\%) the Bayesian test 
  concludes that the two classifiers are practically equivalent ($P(rope)>0.95$)
  despite the significance claimed by the NHST. In this case it draws the opposite conclusion from the NHST.

  Summing up, the Bayesian test with rope 
  is more conservative,   reducing the claimed significances  by $30\%$ as compared with the NHST. 
  The Bayesian test is thus more conservative due to the rope, which constitutes a sensible null hypothesis, while the null hypothesis of the NHST is surely wrong.
  However counting the detection of practically equivalent classifiers  as decisions,
  the Bayesian test with rope takes more decisions 
  than the NHST (222 vs. 199).

  \section{Comparing two classifiers on multiple data sets}
  So far we have discussed how to compare two classifiers on the same data set.
  In machine learning, another important problem is how to compare two classifiers  on a collection of $q$ different data sets, after having 
  performed cross-validation on each data set.
  
  \subsection{The frequentist approach}
  There is no direct NHST able to perform such statistical comparison, i.e., one that takes as inputs the 
  $m$ runs of the $k$-fold cross-validation differences of accuracy for each dataset and returns as output a statistical decision
  about which classifier is better in all the datasets.
  The usual NHST procedure that is employed  for performing such an analysis has two steps:
  \begin{enumerate}
   \item compute  the mean \textit{difference} of accuracy for each dataset (averaging the differences of accuracies
   obtained in the  $m$ runs of the $k$-fold cross-validation);
   \item perform a NHST  to establish if the two classifiers have different performance or not 
   based on these mean differences of accuracy.
  \end{enumerate}
For our case study, \textit{nbc} vs. \textit{aode},  the mean \textit{differences} of accuracy in each dataset 
computed from Table \ref{tab:accuracies} are shown in Table \ref{tab:meanacc}.
We denote these measures generically with $\bm{z}=\{z_1,\dots,z_q\}$ (in our case $q=54$). 
The recommended NHST for this task is the signed-rank test \citep{demvsar2006statistical}.

    \begin{table}[h]
    \rowcolors{1}{white}{lightblue}
  \begin{center}
  \begin{tabular}{rrrrrr}
\text{Dataset}  & \multicolumn{1}{c}{Mean Dif.} & \text{Dataset}  & \multicolumn{1}{c}{Mean Dif.} & \text{Dataset}  & \multicolumn{1}{c}{Mean Dif.}\\\hline
anneal & -1.939 & audiology & -0.261 & breast-cancer & 0.467 \\
cmc & -0.719 & contact-lenses & 2.000 & credit & -0.464 \\
german-credit & -1.014 & pima-diabetes & -0.151 & ecoli & -7.269 \\
eucalyptus & -0.790 & glass & -2.600 & grub-damage & 4.362 \\
haberman & -0.614 & hayes-roth & 0.000 & cleeland-14 & -0.625 \\
hungarian-14 & -0.069 & hepatitis & -0.212 & hypothyroid & -1.683 \\
ionosphere & 0.267 & iris & -3.242 & kr-s-kp & -0.833 \\
labor & 0.000 & lier-disorders & -1.762 & lymphography & -1.863 \\
monks1 & -10.002 & monks3 & -0.343 & monks & -4.190 \\
mushroom & -2.434 & nursery & -4.747 & optdigits & -3.548 \\
page-blocks & 0.583 & pasture & -10.043 & pendigits & -0.443 \\
postoperatie & 1.333 & primary-tumor & -0.674 & segment & -3.922 \\
solar-flare-C & -2.776 & solar-flare-m & -0.688 & solar-flare-X & -3.996 \\
sonar & -0.338 & soybean & -1.112 & spambase & -3.284 \\
spect-reordered & -1.684 & splice & -0.699 & squash-stored & -0.367 \\
squash-unstored & -5.600 & tae & -0.400 & credit & -16.909 \\
owel & -5.040 & waveform & -1.809 & white-clover & 0.500 \\
wine & 0.143 & yeast & -0.202 & zoo & -0.682 \\
  \hline 
  \end{tabular}
  \end{center}
  \caption{Mean difference of accuracy (0--100) for each dataset for nbc minus aode}
  \label{tab:meanacc}
  \end{table}
  
  \begin{SnugshadeF}
\begin{MethodF}[signed-rank test]
The Wilcoxon signed-rank test is a non-parametric statistical hypothesis test used when comparing two paired samples.
 The signed-rank test assumes the observations $z_1,\dots,z_q$ to be i.i.d. and generated from a symmetric distribution.
  The test is miscalibrated if the distribution is not symmetric. A strict usage of the test should thus include first a test for symmetry. One should run the signed-rank
  only if ``the symmetry test fails to reject the null'' (note that, as we discussed before, this does not actually prove that the distribution is symmetric!). However this would make the whole testing procedure cumbersome, requiring also corrections for the test of multiple hypotheses.
  In the common practice thus the test for symmetry is not performed, and we follow this practice in this paper.

  The test is typically used as follows.
  The null hypothesis is that the median of the distribution from which the $z_i$'s are sampled is 0; 
  when the test rejects  the null hypothesis, it claims that it is significantly different from 0.
  The test  ranks the $z_i$'s according to their absolute value and then compares the ranks of the positive differences 
 and negative differences. 
  The test statistic is:
  \begin{equation*} \label{eq:wsignstat}
  \begin{array}{rcl}
  t=&\sum\limits_{\{i:~z_i\geq 0\}} r_i (|z_i|)&= \sum\limits_{1\leq i \leq j \leq q} t^+_{ij},\vspace{2mm} \\
  \end{array}
  \end{equation*} 
  where $r_i (|z_i|)$ is the rank of $|z_i|$ and
  $$
  t^+_{ij}=\left\{\begin{array}{ll}
  1 & \textit{if } z_i \geq -z_j,\\
  0 & \textit{otherwise. } \\
  \end{array}\right.
  $$
  For instance, let us consider the following two cases $\bm{z}=\{-2,-1,4,5\}$ or $\bm{z}=\{-1,4,5\}$, then the statistic is $t=7$
  and, respectively, $t=5$.   For a large enough number of samples (e.g., $q>10$), 
  the statistic  under the null hypothesis is approximately normally distributed and in this case the two-sided test is performed
  as follows:
  \begin{equation}
\begin{array}{l}
w = \frac{t-\frac{q(q+1)}{4}}{\sqrt{\frac{q(q+1)(2q+1) - \text{tie}}{24}}},\vspace{2mm}\\
     p = 2 (1-\Phi(|w|)),
     \end{array}
  \end{equation} 
  where $p$ denotes the $p$-value computed w.r.t.\ $ \Phi$, which is the cumulative distribution function of the standard Normal distribution; $\textit{tie}$ is an adjustment for ties in the data $|\bm{z}|$, i.e., $z_i=-z_j$ for some $i,j$, required by  the nonparametric test \citep{sidak1999,hollander2013nonparametric}, while it is zero in case of no ties.	
  
    Being non-parametric, the signed-rank is robust to outliers.
  It assumes commensurability of differences, but only qualitatively: greater differences  count more as they top the rank; yet their absolute magnitudes are ignored \citep{demvsar2006statistical}.
  \end{MethodF}
  \end{SnugshadeF}

    \subsubsection{Experimental results}
\begin{figure}[h]
    \centering
        \includegraphics[width=7cm]{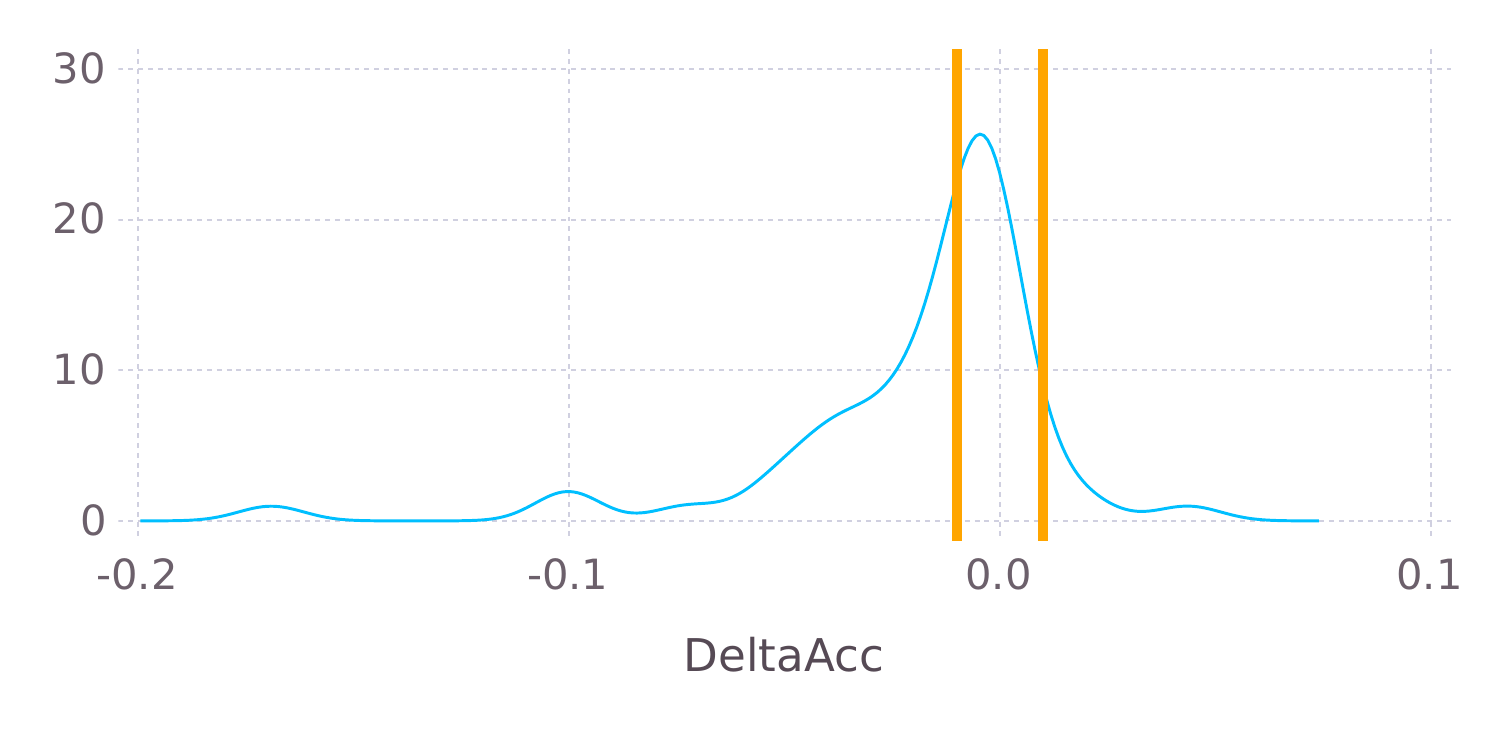}      \includegraphics[width=7cm]{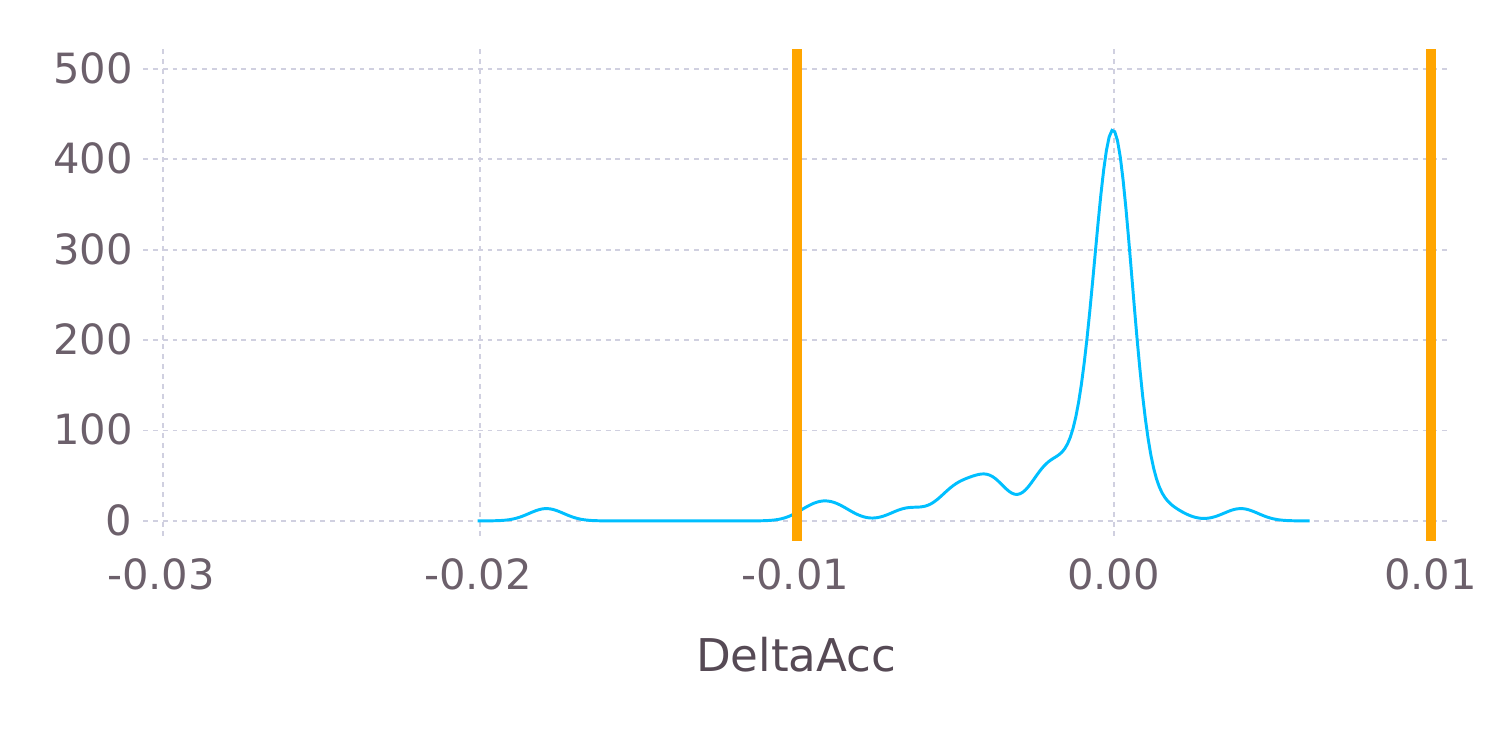} 
        \caption{Density plot for \textit{nbc} versus \textit{aode} (left) and \textit{j48} versus \textit{j48gr} (right)}
                            \label{fig:hist12meanacc}
    \end{figure}

    If we apply this method to compare \textit{nbc} vs.\ \textit{aode}, we obtain
       \text{p-value}=$10^{-6}$ (the rank $t=162$ with no ties and $w$ is $-4.8$). Since the p-value is less than $0.05$, the NHST concludes that the null hypothesis can be rejected 
and that \textit{nbc} and \textit{aode} are significantly different.
        Table \ref{tab:pvaluefive} reports the $p$-values of all comparisons
        of the five classifiers. The pairs \textit{nbc-aode}, \textit{nbc-hnb}, \textit{j48-j48gr} 
        are statistically  significantly different.
                Again by applying this black and white mechanism, we encounter the same problems as before, i.e.,
        we do not have any idea of the magnitude of the effect size, the uncertainty, 
        the probability of the null hypothesis, et cetera.
        The density plot of the data (the mean differences of accuracy in each dataset) for \textit{nbc} versus \textit{aode} shows for instance that there are many datasets
        where the mean difference is small (close to zero), see Figure \ref{fig:hist12meanacc}.
        Instead for \textit{j48} versus \textit{j48gr}, it is clear that the difference of accuracy is very small. 
        
            \begin{table}[h]
    \rowcolors{1}{white}{lightorange}
  \begin{center}
  \begin{tabular}{ccc}
\text{Classif. 1}  & \multicolumn{1}{c}{Classif. 2} & \text{$p$-value} \\\hline
nbc & aode & 0.000 \\
nbc & hnb & 0.001 \\
nbc & j48 & 0.463 \\
nbc & j48gr & 0.394 \\
aode & hnb & 0.654 \\
aode & j48 & 0.077 \\
aode & j48gr & 0.106 \\
hnb & j48 & 0.067 \\
hnb & j48gr & 0.084 \\
j48 & j48r & 0.000\\
  \hline 
  \end{tabular}
  \end{center}
  \caption{$p$-values for the comparison of the five classifiers.}
  \label{tab:pvaluefive}
  \end{table}

  \subsection{The Bayesian analysis approach}
  We will present two ways of approaching the comparison  between two classifiers in multiple datasets.
  The first will be based on a nonparametric approach that directly extends the Wilcoxon signed-rank test.
  The second is a hierarchical model.

  \subsubsection{Nonparametric test}
\cite{benavoli2014a} have proposed a Bayesian counterpart of the frequentist  sign and signed-rank test,
which is based on the Dirichlet process.
    \begin{SnugshadeB}
\begin{MethodB}[sign and signed-tank test]
Let $z$ denote the scalar variable of interest and $\bm{z}=\{z_1,\dots,z_q\}$ denotes a vector
of i.i.d.\ observations of $z$.
To derive the Bayesian sign and signed-rank test, we assume a Dirichlet Process (DP)
prior on the probability distribution of  $z$. 
A DP is a distribution over probability distributions such that marginals 
on finite partitions are Dirichlet distributed.
Like the Dirichlet distribution, the DP is therefore completely defined by two parameters:  the prior strength $s>0$
and the prior mean that, for the DP, is a probability measure $G_0$ on $z$.
If we choose $G_0=\delta_{z_{0}}$, i.e., a Dirac's delta centred on the pseudo-observation $z_0$, 
the posterior probability density function of $Z$ has this simple expression:

\begin{equation}
\label{eq:postDP}
p(z)=w_{0}\delta_{z_{0}}(z)+\sum_{j=1}^{n} w_{j} \delta_{z_j}(z), \;\;\; (w_{0},w_{1},\dots,w_{n})\sim Dir(s,1,\dots,1),
\end{equation} 
i.e., it is a mixture of Dirac's deltas centred on the observations $z_j$ for $j=1,\dots,q$
and on the prior pseudo-observation $z_0$, whose
weights are Dirichlet distributed with parameters $(s,1,\dots,1)$.
We can think about (\ref{eq:postDP}) as a hierarchical model: $p(z)$ depends on
the weights $w$ that are Dirichlet distributed.
The model (\ref{eq:postDP}) is therefore a posterior distribution of the probability 
distribution of $z$ and encloses all the information we need for the experimental analysis.
We can summarize it in different way.
If we compute:
 \begin{align}
\nonumber
 \theta_{l}&=P(z<-r)=\sum_{i=0}^q w_i I_{(-\infty,-r)}(z_i), \\
\nonumber
 \theta_{e}&=P(|z|\leq r)=\sum_{i=0}^q w_i I_{[-r,r]}(z_i),\\
\nonumber
 \theta_{r}&=P(z>r)=\sum_{i=0}^q w_i I_{(r,\infty)}(z_i),
\end{align}
where the indicator $I_{A}(z)=1$ if $z_0 \in A$  and zero otherwise, then we obtain a Bayesian version of the \textbf{sign test} that also accounts for the rope $[-r,r]$.
In fact, $ \theta_{l}, \theta_{e}, \theta_{r}$ are respectively the probabilities that the mean difference of accuracy 
is in the interval $(-\infty,-r)$, $[-r,r]$, or $(r,\infty)$.
Since $(w_{0},w_{1},\dots,w_{n})\sim Dir(s,1,\dots,1)$, it can easily be shown that
\begin{equation}
\label{eq:signtestrope}
\theta_l,\theta_e,\theta_r \sim Dirichlet(n_{l}+sI_{(-\infty,-r]}(z_0),n_{e}+sI_{[-r,r]}(z_0),n_{r}+sI_{[r,\infty)}(z_0)),
\end{equation}
where $n_l$ is the number of observations $z_i$ that fall in $(-\infty,-r]$, $n_e$ is the number of observations $z_i$ that fall in $[-r,r]$
and $n_r$ is the number of observations $z_i$ that fall in $[r,\infty)$, obviously $n_l+n_e+n_r=q$. If we neglect $sI_{(-\infty,-r]}(z_0),sI_{[-r,r]}(z_0),sI_{[r,\infty)}(z_0)$, then (\ref{eq:signtestrope}) says that the posterior probability
of $\theta_l,\theta_e,\theta_r$ is Dirichlet distributed with parameters $(n_l,n_e,n_r)$.
The terms $sI_{(-\infty,-r]}(z_0),sI_{[-r,r]}(z_0),sI_{[r,\infty)}(z_0)$ are due to the prior. Therefore, to fully specify the Bayesian sign test, we must choose
the value of the prior strength $s$ and where to place the pseudo-observation $z_0$ in $(-\infty,-r]$ or $[-r,r]$ or $[r,\infty)$.
We will return to this choice in Section \ref{sec:prior}.
%
Instead, if we compute 
 \begin{align}
 \nonumber
\theta_{l}&=\sum_{i=0}^q \sum_{j=0}^q w_iw_j I_{(-\infty,-2r)}(z_j+z_i),\\
\nonumber
\theta_{e}&=\sum_{i=0}^q \sum_{j=0}^q w_iw_j I_{[-2r,2r]}(z_j+z_i),\\
 \label{eq:signrankp}
 \theta_{r}&=\sum_{i=0}^q \sum_{j=0}^q w_iw_j I_{(2r,\infty)}(z_j+z_i),
 \end{align}
then we derive a Bayesian version of the \textbf{signed rank test} \citep{benavoli2014a} 
that also accounts for the rope $[-r,r]$. 
This time the distribution of  $\theta_l,\theta_e,\theta_r$ has not a simple closed form but we can easily compute
it by Monte Carlo sampling the weights $(w_{0},w_{1},\dots,w_{n})\sim Dir(s,1,\dots,1)$.
Also in this case we must choose $s,z_0$, see Section \ref{sec:prior}.
 \end{MethodB}
\end{SnugshadeB}
It should be observed that the Bayesian signed-rank test does not require the symmetry assumption
about the distribution of the observations $z_i$. This test works also in case the distribution is asymmetric
thanks to the Bayesian estimation approach (i.e., it estimates the distribution from data).
This is another advantage of the Bayesian estimation approach w.r.t.\ the frequentist null hypothesis tests \citep{benavoli2014a}.

\subsubsection{Experiments}
\begin{figure}[!h]
    \centering
        \includegraphics[width=7cm]{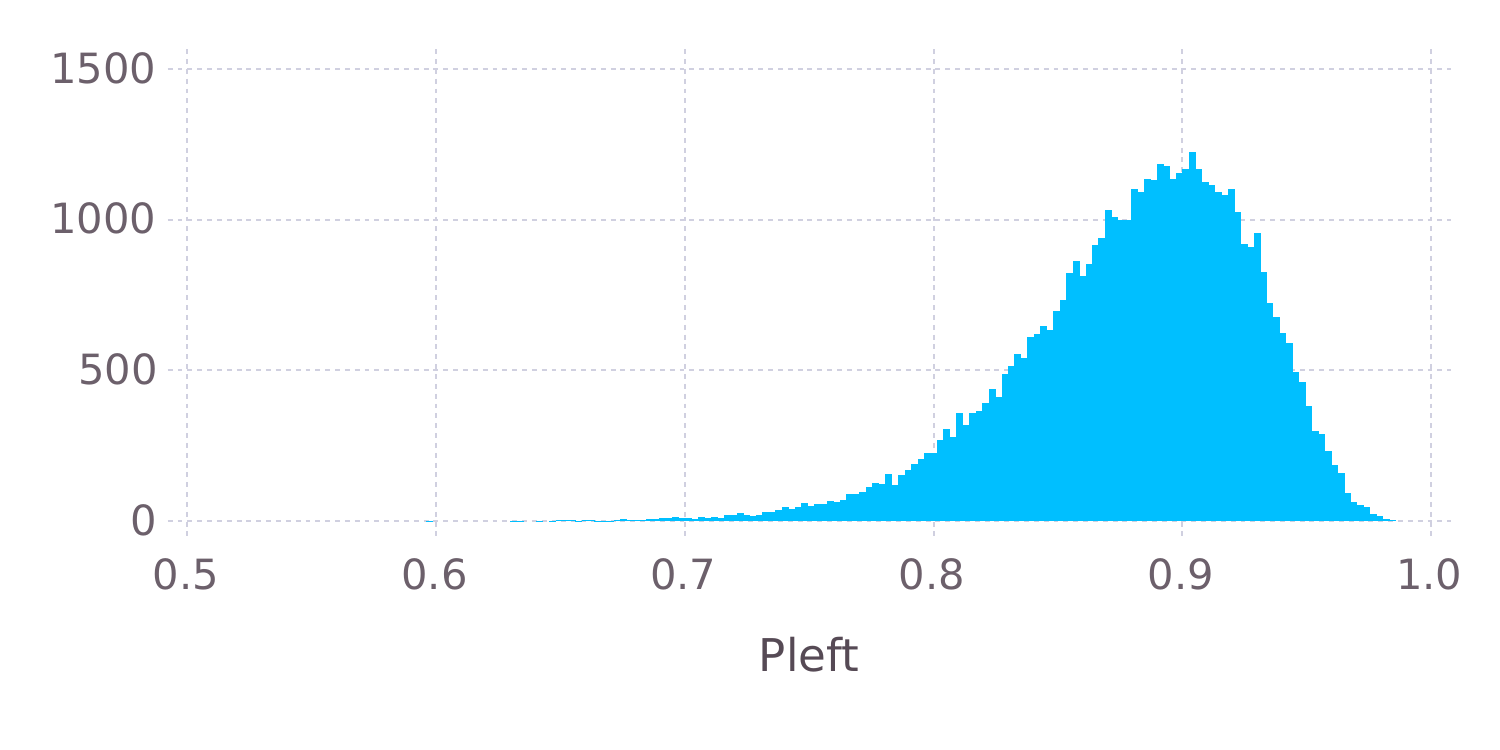}
        \caption{Posterior for  \textit{nbc} vs.\ \textit{aode} for the Bayesian sign-rank test.}
                            \label{fig:BAsignborope}
    \end{figure}

Let  us start by comparing \textit{nbc} vs.\ \textit{aode} by means of
the Bayesian sign-rank tests \textit{without} rope ($r=0$).
Hereafter we will choose the prior parameter of the Dirichlet  as  $s=0.5$ and $z_0=0$; we will return to this choice in Section \ref{sec:prior}.
Since without rope $\theta_r=1-\theta_l$, we have only reported the posterior
of $\theta_l$ (denoted as ``Pleft'') that represents the probability that \textit{aode} is better than \textit{nbc}.
The samples of the posteriors  are shown in  Figure \ref{fig:BAsignborope}: this is simply the histogram of $150'000$ 
samples of $\theta_l$ generated according to \eqref{eq:signrankp}.
For all samples, it results in $\theta_l$ greater than $0.5$ and so  $\theta_r=1-\theta_l$. So we can conclude with probability $\approx 1$
that \textit{aode} is better than \textit{nbc}. We can in fact think about the comparison of two classifiers as the inference
on the bias ($\theta_l$) of a coin. In this case, all the $150'000$ sampled coins from the posterior have a bias that is greater than $0.5$ and, therefore,
all the coins are always biased towards  \textit{aode} (which is then preferable to  \textit{nbc}).

This conclusion is in agreement with that derived by the frequentist  sign-rank test (very small $p$-value, see Table \ref{tab:pvaluefive}). 
         
         \begin{figure}[!h]
         	\centering
         	\includegraphics[width=7cm]{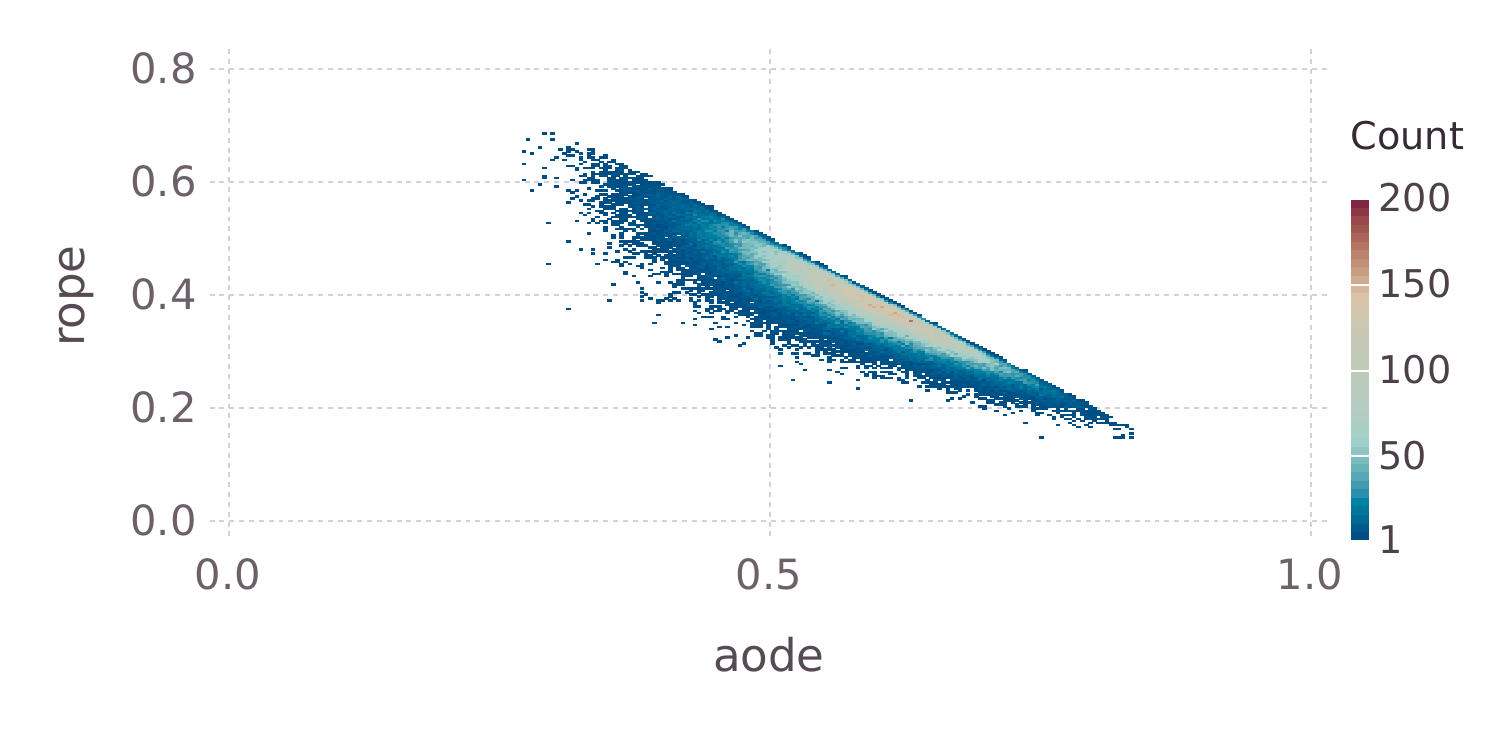}         \includegraphics[width=7cm]{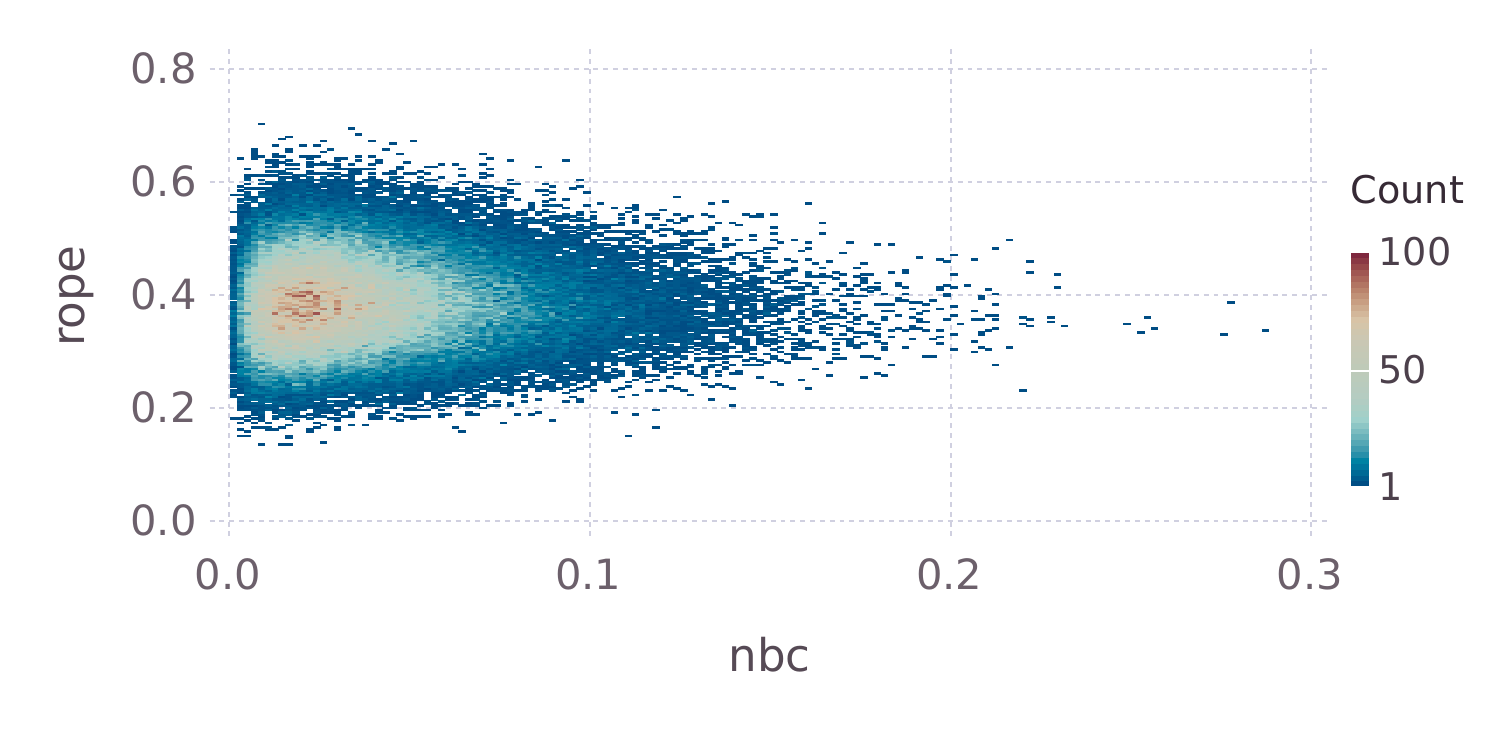}
         	\includegraphics[width=7cm]{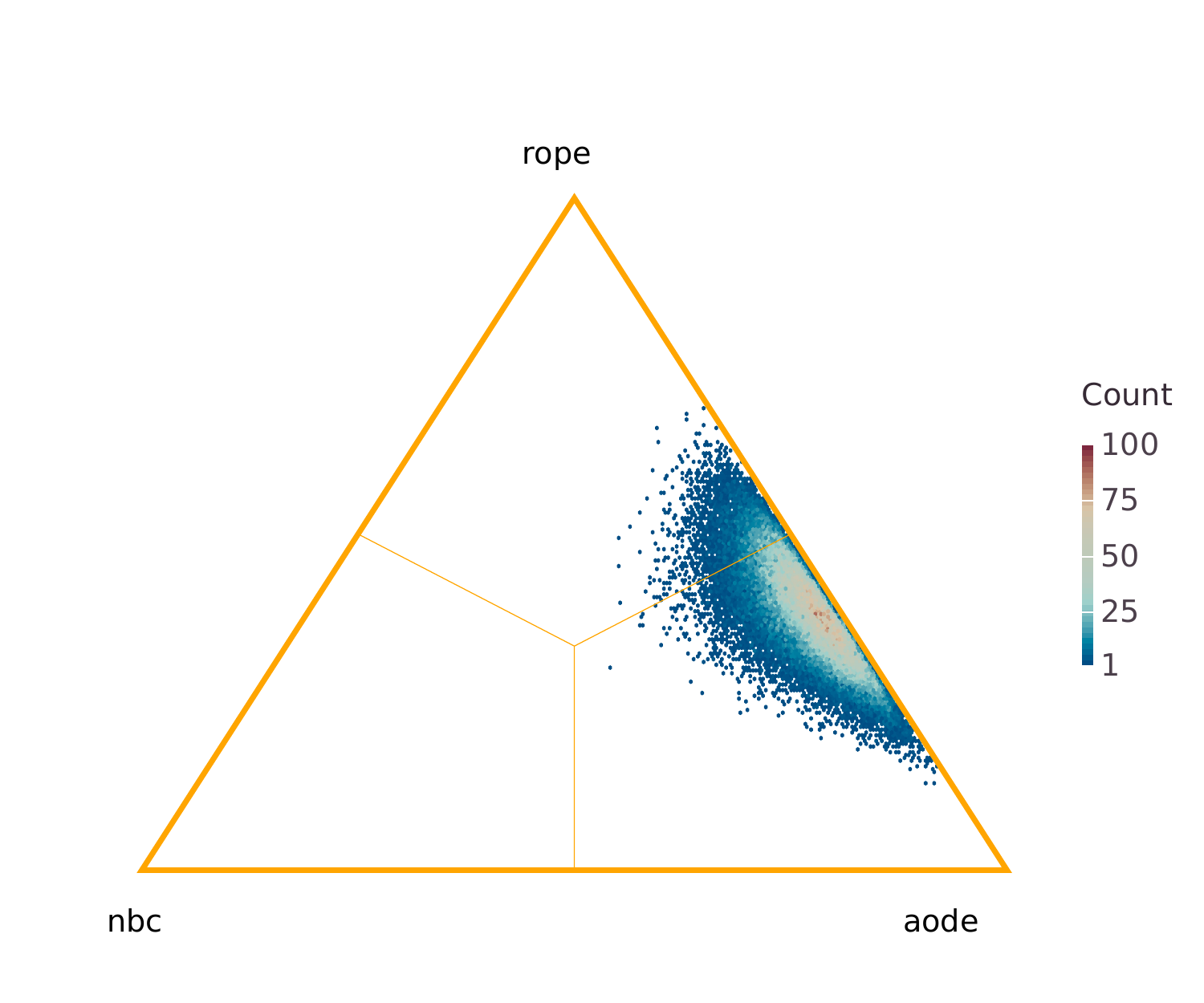}
         	\caption{Posterior for  \textit{nbc} vs.\ \textit{aode} for the Bayesian sign-rank test.}
         	\label{fig:BAsignrope2}
         \end{figure}

         \begin{figure}[!h]
         	\centering
         	
         	\includegraphics[width=15cm]{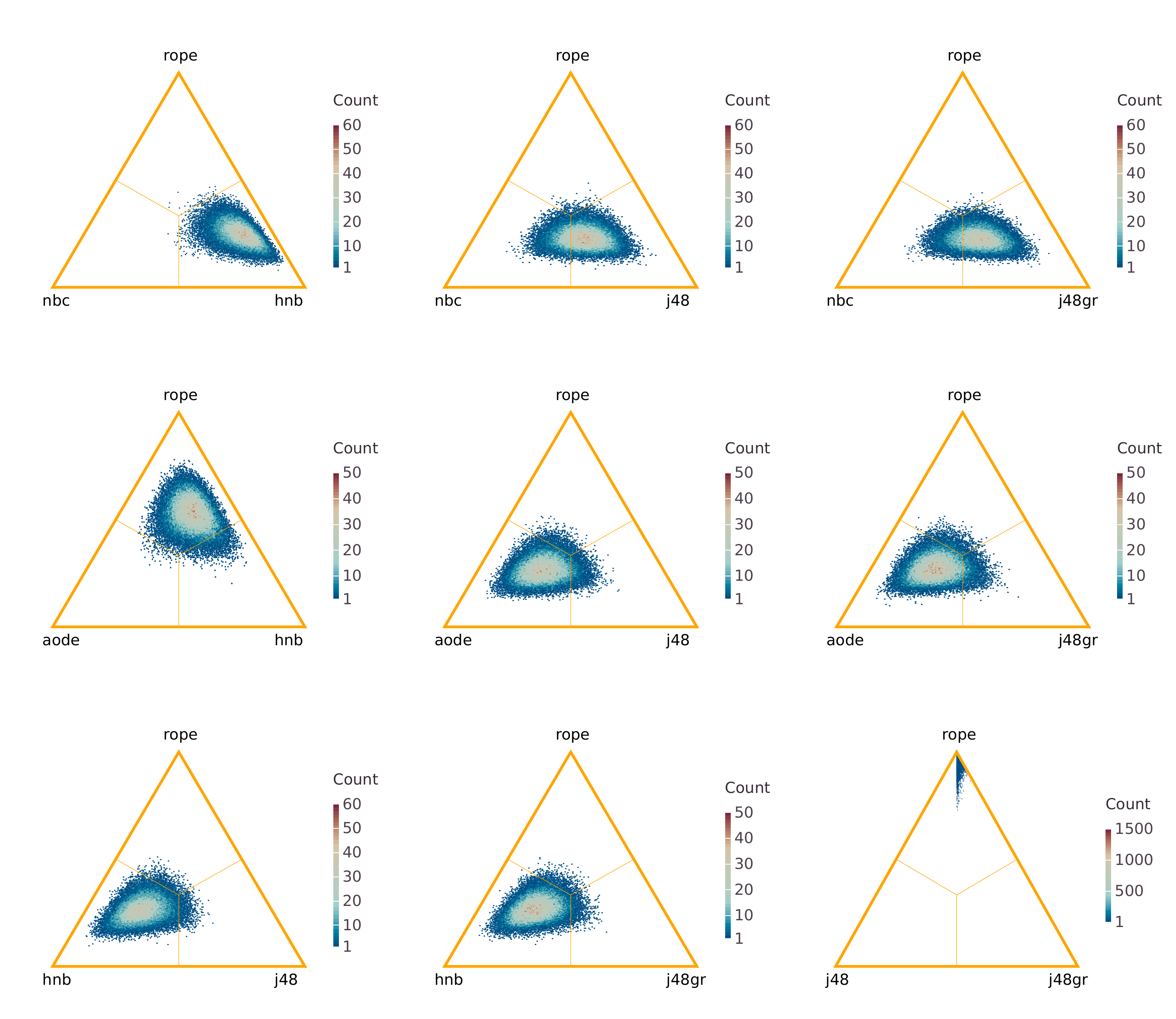}
         	\caption{Posterior for  \textit{nbc} vs.\ \textit{aode} for Bayesian sign-rank test.}
         	\label{fig:plotmanytriangles}
         \end{figure}
    The introduction of the \textit{rope} partially changes the previous conclusion.
    A way to visualize the posterior of $\theta_l,\theta_e,\theta_r$ in this case, is by plotting
    the $150'000$ Monte Carlo samples of these probabilities in barycentric coordinates: each 
    trinomial vector of probabilities is a point in the simplex having vertices $\{(1,0,0),(0,1,0),(0,0,1)\}$.
    The three vertices are respectively denoted as ``aode'', ``rope'' and ``nbc'' and represent
    decisions with certainty in favour of ``aode'', ``rope'' and, respectively, ``nbc''.
    Figure \ref{fig:BAsignrope2}  reports the simplex as well as the two-dimensional projections
    of the posterior for the Bayesian  sign-rank test.
    In particular Figure \ref{fig:BAsignrope2} reports the marginal of the posterior distribution
    of  ``aode'' vs.\ ``rope'' (left); the marginal of the posterior distribution
    ``nbc'' vs.\ ``rope'' (right). From these two figures we can deduce the other marginal since
    $\theta_l+\theta_e+\theta_r=1$. Finally, Figure~\ref{fig:BAsignrope2}~(bottom) reports the joint of the three variables
    in barycentric coordinates (we are again exploiting the fact that  $\theta_l+\theta_e+\theta_r=1$).
    In particular, Figure \ref{fig:BAsignrope2} (bottom) reports the samples from the posteriors (cloud of points), the simplex (the large orange triangle)
    and three regions (in orange) that are limited by the level curves: $\theta_i\geq \max(\theta_j,\theta_k)$ with $i\neq j \neq k$ (hypothesis $i$
    is more probable than both hypotheses $j,k$ together).
    For instance, the region at the bottom-right of the triangle is relative to the case where ``aode'' is more probable
    than ``rope'' and ``nbc'' together; the region at the top of the triangle represents  the case where ``rope'' is more probable     than ``aode'' and ``nbc'' together; the region at the left of the triangle corresponds to the case where ``nbc'' is more probable     than ``aode'' and ``rope'' together. 
    Hence, if all the points fall inside one of these three regions, we conclude that such hypothesis is true with probability $\approx 1$.
    Looking at Figure \ref{fig:BAsignrope2} (bottom), it is evident that the majority of cases support 
    \textit{aode} more than \textit{rope} and definitively more than \textit{nbc}.
    We can quantify this numerically by counting the number of points that fall in the three regions, see first row in Table \ref{tab:signranktestprob}.
    \textit{aode} is better in $90\%$ of cases, while \textit{rope} is selected in the remaining $10\%$.
    We can therefore conclude with probability $90\%$ that \textit{aode} is practically better than \textit{nbc}.
    Table \ref{tab:signranktestprob} reports also these probabilities for the other comparisons of classifiers computed
    using $150'000$ Monte Carlo samples.
    We conclude that \textit{hnb} is practically better than \textit{nbc} with probability $0.999$;
    \textit{aode} and \textit{hnb} are equivalent with probability $0.95$; \textit{aode} is better than
    \textit{j48} and \textit{j48gr} with probability $0.9$; \textit{hnb} is better than  \textit{j48} and \textit{j48gr} with probability greater than $0.95$
    and finally \textit{j48} and \textit{j48gr} are practically equivalent.
   These conclusions are in agreement with the data.
   
   The computational complexity of the Bayesian sign-rank test is low.
   The comparison of two classifiers (based on  $150'000$ samples) takes less than one second on a standard computer.

        \begin{table}[!h]
    \rowcolors{1}{white}{lightblue}
  \begin{center}
  \begin{tabular}{rrrrr}
\text{Classif. 1}  & \multicolumn{1}{c}{Classif. 2} & \text{left} &  \text{rope}  & \text{right}\\\hline
nbc & aode & 0.000 & 0.103 & 0.897 \\
nbc & hnb & 0.000 & 0.001 & 0.999 \\
nbc & j48 & 0.228 & 0.004 & 0.768 \\
nbc & j48gr & 0.182 & 0.002 & 0.815 \\
aode & hnb & 0.001 & 0.956 & 0.042 \\
aode & j48 & 0.911 & 0.026 & 0.063 \\
aode & j48gr & 0.892 & 0.035 & 0.073 \\
hnb & j48 & 0.966 & 0.015 & 0.019 \\
hnb & j48gr & 0.955 & 0.020 & 0.025 \\
j48 & j48gr & 0.000 & 1.000 & 0.000 \\
  \hline 
  \end{tabular}
  \end{center}
  \caption{Probabilities for the ten comparisons of classifiers. Left and right refer to the columns Classif. 1 (left)
  and Classif. 2 (right).}
  \label{tab:signranktestprob}
  \end{table}

\subsection{Choice of the prior}
\label{sec:prior}
In the previous section, we have selected the prior parameters of the Dirichlet process as $s=0.5$ and $z_0=0$.
In terms of rank, this basically means that the prior strength is equivalent to that of one pseudo-observation that is located inside the rope
\citep{benavoli2014a}.
How are the inferences sensitive to this choice? For instance, we can see how the probabilities 
on Table \ref{tab:signranktestprob} would change based on $z_0$. We have considered two extreme cases $z_0=-\infty$ and $z_0=\infty$
and reported these probabilities in Table \ref{tab:signranktestprob1} and \ref{tab:signranktestprob2} (this is an example of robust Bayesian  analysis 
\citep{r1994orb}).
It is evident that the  position of $z_0$ has only a minor effect on the probabilities.
We could have performed this analysis jointly by considering all the possible Dirichlet process priors
obtained by varying $z_0\in \mathbb{R}$. This set of Dirichlet priors  is called ``Imprecise Dirichlet Process'' (IDP).
IDP  allows us to start the inference  with very weak prior assumptions, much in the direction of letting data speak for themselves. 
More details about the properties of IDP and the choice of the prior can be found in \cite{benavoli2014a,benavoli2014b,walley1996}.

        \begin{table}[!h]
    \rowcolors{1}{white}{lightblue}
  \begin{center}
  \begin{tabular}{rrrrr}
\text{Classif. 1}  & \multicolumn{1}{c}{Classif. 2} & \text{left} &  \text{rope}  & \text{right}\\\hline
nbc & aode & 0.000 & 0.112 & 0.888 \\
nbc & hnb & 0.000 & 0.001 & 0.999 \\
nbc & j48 & 0.262 & 0.004 & 0.734 \\
nbc & j48gr & 0.213 & 0.003 & 0.784 \\
aode & hnb & 0.002 & 0.961 & 0.037 \\
aode & j48 & 0.922 & 0.024 & 0.053 \\
aode & j48gr & 0.906 & 0.033 & 0.061 \\
hnb & j48 & 0.971 & 0.014 & 0.016 \\
hnb & j48gr & 0.961 & 0.018 & 0.021 \\
j48 & j48gr & 0.000 & 1.000 & 0.000 \\
  \hline 
  \end{tabular}
  \end{center}
  \caption{Probabilities for the ten comparisons of classifiers  with  $z_0=\infty$. Left and right refer to the columns Classif. 1 (left)
  and Classif. 2 (right).}
  \label{tab:signranktestprob1}
  \end{table}
  
          \begin{table}[!h]
    \rowcolors{1}{white}{lightblue}
  \begin{center}
  \begin{tabular}{rrrrr}
\text{Classif. 1}  & \multicolumn{1}{c}{Classif. 2} & \text{left} &  \text{rope}  & \text{right}\\\hline
nbc & aode & 0.000 & 0.096 & 0.904 \\
nbc & hnb & 0.000 & 0.001 & 0.999 \\
nbc & j48 & 0.201 & 0.004 & 0.795 \\
nbc & j48gr & 0.159 & 0.002 & 0.839 \\
aode & hnb & 0.001 & 0.950 & 0.049 \\
aode & j48 & 0.892 & 0.028 & 0.080 \\
aode & j48gr & 0.872 & 0.037 & 0.091 \\
hnb & j48 & 0.957 & 0.017 & 0.027 \\
hnb & j48gr & 0.944 & 0.022 & 0.034 \\
j48 & j48gr & 0.000 & 1.000 & 0.000 \\
  \hline 
  \end{tabular}
  \end{center}
  \caption{Probabilities for the ten comparisons of classifiers  with  $z_0=-\infty$. Left and right refer to the columns Classif. 1 (left)
  and Classif. 2 (right).}
  \label{tab:signranktestprob2}
  \end{table}

  \subsubsection{Hierarchical models}
  In Section \ref{sec:student} we have presented the Bayesian correlated t-test  that is used for the analysis of cross-validation results on a single dataset.
  In particular, it makes inference about the mean difference of accuracy between two classifiers in the $i$-th dataset ($\mu_i$)  by exploiting three pieces of information: the sample mean ($\bar{x}_i$), the variability of the data (sample standard deviation $\hat{\sigma}_i$) and the correlation due to the overlapping training set ($\rho$). This test can only be applied to a single dataset. 
We have already discussed the fact that there is no direct NHST able to extend the above  statistical comparison to multiple datasets, i.e., that takes as inputs the   $m$ runs of the $k$-fold cross-validation results for each dataset and returns as output a statistical decision
  about which classifier is better in all the datasets.
  The usual NHST procedure that is employed  for performing such analysis has two steps:
 (1) compute  the mean \textit{difference} of accuracy for each dataset $\bar{x}_i$;
   (2) perform a NHST  to establish if the two classifiers have different performance or not 
   based on these mean differences of accuracy.
This discards two pieces of information: the correlation $\rho$ and sample standard deviation $\hat{\sigma}_i$ in each dataset.
The standard deviation is informative about  the accuracy of $\bar{x}_i$  as an estimator of  $\mu_i$.
 The standard deviation can largely vary  across data sets, as a result of each data set having its own size and complexity.
 The aim of this section is to present an extension  of the  Bayesian correlated t-test that is able
 to make inference on multiple datasets and at the same time  to account for all the available information
 (mean, standard deviation and correlation).
 In Bayesian estimation, this can be obtained by defining a hierarchical model \citep{corani2016unpub}.
 Hierarchical models are among the most powerful and flexible tools in Bayesian analysis.

      \begin{SnugshadeB}
\begin{MethodB}[hierarchical correlated t-test]
The  hierarchical correlated t-test is based on following hierarchical probabilistic model:
  \begin{align}
  	&	\mathbf{x}_{i} \sim  MVN(\mathbf{1} \mu_i,\mathbf{\Sigma_i})  \label{eq:mvn}, \\
  	&	\mu_1 ... \mu_q \sim t (\mu_0, \sigma_0,\nu ) \label{eq:delta_i}, \\
  	& \sigma_1 ... \sigma_q \sim \mathrm{unif} (0,\bar{\sigma}). \label{eq:sigma_i} 
  \end{align}
  Equation (\ref{eq:mvn}) models the fact that the cross-validation measures $\bm{x_i}=\{x_{i1},x_{i2},\dots,x_{in}\}$  of the i-th data set are jointly multivariate-normal distributed with the same mean ($ \mu_i$), same variance ($\sigma_i$) and correlation ($\rho$).
  In fact, it states that, for each dataset $i$,  $\mathbf{x}_{i}$  is multivariate normal with mean $\mathbf{1} \mu_i$ (where $\mathbf{1}$ is a vector of ones)  and
  covariance matrix $\mathbf{\Sigma}_i$ defined as follows: the diagonal elements are $\sigma^2_i$ and the out-of-diagonal elements are
  $\rho\sigma^2_i$, where $\rho=\frac{n_{te}}{n_{tr}}$.      This model is the same we discussed in Section \ref{sec:student}.
  Equation (\ref{eq:delta_i}) models the fact that the mean difference of accuracies 
  in the single datasets, $\mu_i$, depends on $\mu_0$ that is the average difference of accuracy between the two classifiers on the population 
  of data sets. This is the quantity we aim at estimating.
  Equation (\ref{eq:delta_i})  assumes the $\mu_i$'s to be drawn from a high-level Student distribution
  with mean  $\mu_0$, variance  $\sigma^2_0$ and degrees of freedom $\nu$.
  The choice of a Student distribution at this level of the hierarchical model enables 
  the  model to robustly deal with data sets whose $\mu_i$'s are far away from the others  \citep{gelman2014bayesian,kruschke2013bayesian}.
  Moreover the heavy tails of the Student make more cautious  the conclusions drawn by the model.
  
  The hierarchical model    assigns to the i-th data set its own standard deviation $\sigma_i$, assuming the $\sigma_i$'s to be drawn from a common distribution,   see   Equation \eqref{eq:sigma_i}.
  In this way it realistically represents the fact the estimates   referring to different data sets
  data sets have different uncertainty.     The high-level distribution of the $\sigma_i$'s is $\mathrm{unif}(0,\bar{\sigma})$, as recommended by \cite{gelman2006prior}, as it yields inferences which are insensitive to $\bar{\sigma}$, if $\bar{\sigma}$ is large enough. To this end we set $\bar{\sigma}= 1000\cdot\bar{s}$ \citep{kruschke2013bayesian}, where $\bar{s}=\sum_i^q \hat{\sigma}_i/q$.
  
  We complete the model with the prior on the parameters 
$\delta_0$, $\sigma_0$ and $\nu$
of the high-level distribution.
We assume  $\delta_0$ 
to be uniformly distributed
within 1 and -1. This choice works for 
all the measures  bounded within $\pm$1, such as 
accuracy, AUC, precision and recall. Other type of indicators might require different bounds.

For the standard deviation
$\sigma_0$ we adopt the prior  $unif(0,\bar{s_0})$, with
$\bar{s_0}=1000 s_{\bar{x}}$, where $s_{\bar{x}}$ is the  standard deviation of the 
$\bar{x}_i$'s. 

As for the prior $p(\nu)$ on the 
degrees of freedom, there are two proposals in the literature.  \cite{kruschke2013bayesian} 
proposes an exponentially shaped distribution which balances  the prior probability of nearly normal distributions ($\nu$ \textgreater 30) and heavy tailed distributions ($\nu$ \textless 30).
We re-parameterize this distribution as a 
Gamma($\alpha$,$\beta$) with
$\alpha$=1, $\beta$= 0.0345.
\cite{juarez2010model} proposes instead 
$p(\nu) = \mathrm{Gamma}(2,0.1)$, assigning larger prior probability to normal distributions.

We have no reason for preferring a prior over another, 
but  the hierarchical model shows some sensitivity on the choice of $p(\nu)$.
We model this uncertainty by representing  the coefficients $\alpha$ and $\beta$ 
of the Gamma distribution as two random variables (hierarchical prior).
In particular we assume
$p(\nu)=\mathrm{Gamma}(\alpha,\beta)$, with 
$\alpha \sim \mathrm{unif} (\underline{\alpha},\bar{\alpha})$ and $\beta \sim \mathrm{unif} (\underline{\beta}, \bar{\beta})$, setting 
$\underline{\alpha}$=0.5, $\bar{\alpha}$=5, $\underline{\beta}$=0.05, $\bar{\beta}$=0.15.
The simulations in \citep{corani2016unpub} show that the inferences of the model are stable with respect to perturbations of 
$\underline{\alpha}$, $\bar{\alpha}$, $\underline{\beta}$, and $\bar{\beta}$, and that the resulting hierarchical generally fits well the experimental data.

  
  These considerations are reflected by the following probabilistic model:
  \begin{align}
  	& \nu \sim Ga(\alpha,\beta), \\
  	& \alpha \sim \mathrm{unif}(\underline{\alpha},\overline{\alpha}), \\
  	& \beta \sim \mathrm{unif}(\underline{\beta},\overline{\beta}), \\
  	& \mu_0 \sim \mathrm{unif}(-1,1),  \\
  	& \sigma_0 \sim \mathrm{unif}(0,\bar{\sigma_{0}}). 
  \end{align}
We want to make inference about the $\mu_i$'s and $\mu_0$.
Such inferences are computed by marginalizing out the $\sigma_i$'s,
 and thus accounting for the different uncertainty which characterizes each data set.
This characteristic is unique among the methods discussed so far.
Computations in hierarchical models are obtained by Markov-Chain Monte Carlo sampling.
\end{MethodB}
\end{SnugshadeB}
                    A further merit of the hierarchical model is that it jointly estimates the   $\mu_i$'s 
                    while
                    the existing  methods estimate independently the difference of accuracy on each data set using the $\bar{x}_i$'s.
                    The consequence of the joint estimation performed by the hierarchical model is that
                    \textit{shrinkage} is applied to  the $\bar{x}_i$'s.
                    The  hierarchical model thus estimates the   $\mu_i$'s more accurately
                    than the $\bar{x}_i$'s adopted by the other tests.
                    This result is valid under general assumptions, such as a severe 
                    misspecification between the high-level distributions of the true generative model and of the fitted model \citep{corani2016unpub}.
                     By applying the rope on the posterior distribution of  the  $\mu_i$'s and the $\mu_0$ in a similar way to
                    what discussed for the Bayesian correlated t-test, the model is able to detect equivalent classifiers and to claim significances that have a practical impact.

\subsubsection{Experiments}
\begin{figure}[h]
    \centering
        \includegraphics[width=7cm]{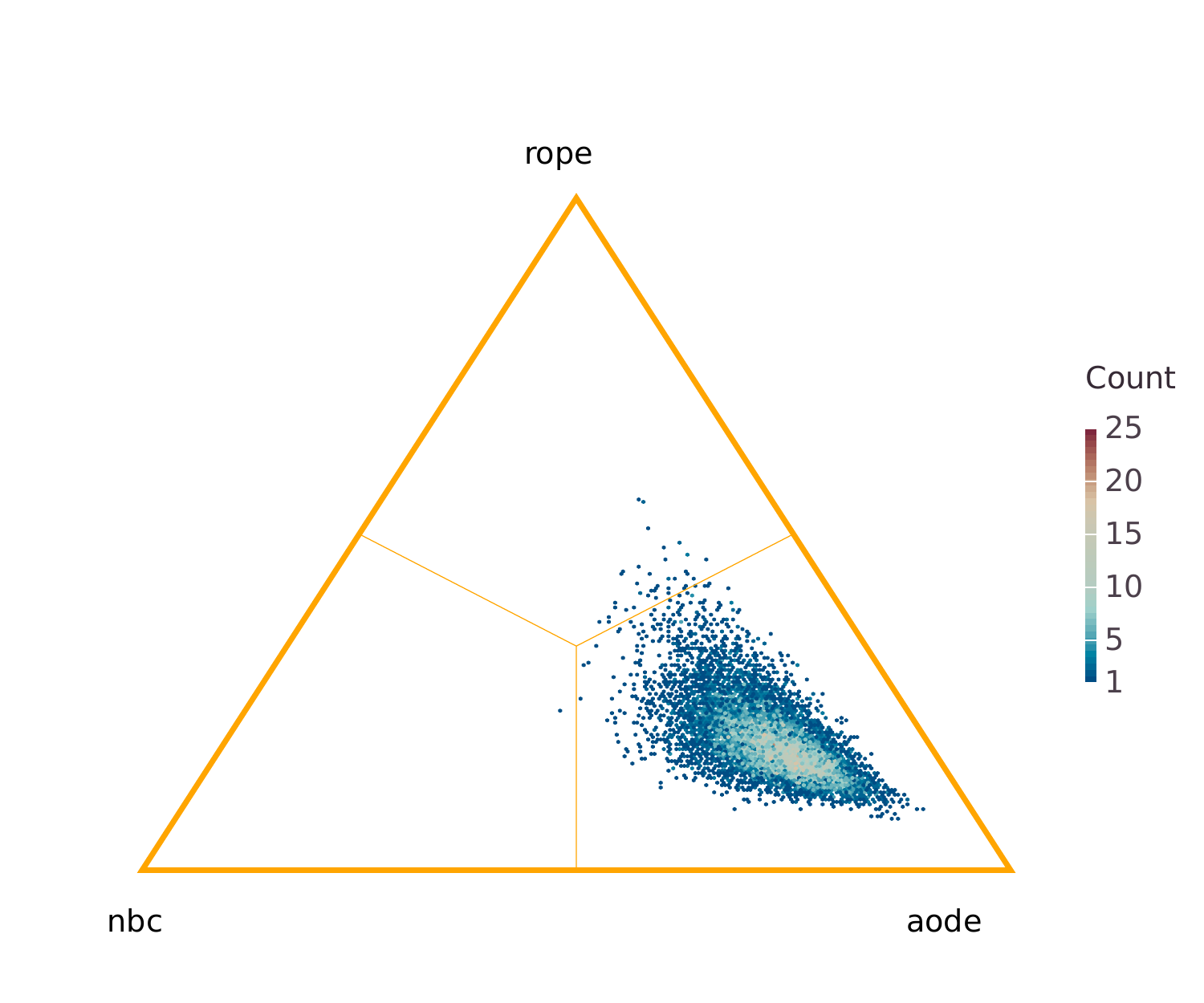}        
        \caption{Posterior of \textit{nbc} versus \textit{aode} for all $54$ datasets.}
                    \label{fig:hierarchipost}
    \end{figure}

In the experiments, we have computed the posterior of $\mu_0, \sigma_0,\nu $
for the ten pairwise comparisons between the classifiers  \textit{nbc}, \textit{aode},
\textit{hnb}, \textit{j48} and \textit{j48gr}.
As inference we have computed the prediction on the next (unseen) dataset, which is formally equivalent to the inference
computed by the Bayesian signed-rank test.
For instance, for  \textit{nbc} vs. \textit{aode}, we have computed the probabilities that in the next dataset
 \textit{nbc} is better than \textit{aode} ($\theta_r$), \textit{nbc} is equivalent to \textit{aode} ($\theta_e$),
 \textit{aode} is better than \textit{nbc} ($\theta_l$).
 This is the procedure we have followed:
 \begin{enumerate}
  \item we have sampled $\mu_0, \sigma_0,\nu $ from the posteriors of these parameters;
  \item for each sample of $\mu_0, \sigma_0,\nu $ we have defined the posterior of the mean difference of accuracy on the next dataset, i.e., $t(\mu_{next};\mu_0, \sigma_0,\nu)$;
  \item from $t(\mu_{next};\mu_0, \sigma_0,\nu)$ we have computed the probabilities $\theta_l$ (integral on $(-\infty,r])$, $\theta_e$ (integral on $[-r,r]$) and $\theta_r$ (integral on $[r,\infty))$.
 \end{enumerate}
We have repeated this procedure  $4'000$ times, obtaining $4'000$ samples of $(\theta_l,\theta_e,\theta_r)$
and the results are shown in Figure \ref{fig:hierarchipost}.
The results are quite in agreement with those of the Bayesian signed-rank test.
For instance, it is evident that \textit{aode}  is  clearly better than \textit{nbc}. 
 We can quantify this numerically by counting the number of points that fall in the three regions (see the first row in Table \ref{tab:hierarchicalres}).
    \textit{aode} is better in  almost $100\%$ of cases.
    Table \ref{tab:hierarchicalres} reports also these probabilities for the other comparisons of classifiers computed
    using $4'000$ Monte Carlo samples.
By comparing Tables \ref{tab:hierarchicalres} and \ref{tab:signranktestprob}, we can see that the two
tests are substantially in agreement apart from  differences in \textit{aode} vs. \textit{j48} and \textit{j48gr}.
The hierarchical test is taking into account of all available infromation: the sample mean ($\bar{x}_i$), the variability of the data (sample standard deviation $\hat{\sigma}_i$) and the correlation due to the overlapping training set ($\rho$), while the Bayesian signed rank only considers  $\bar{x}_i$. Therefore, when the two tests differ substantially,
it means that there is substantial variability  of the cross-validation estimate. 

  We have implemented the hierarchical model in Stan (\url{http://mc-stan.org}) \citep{carpenterstan}, a language for Bayesian inference. 
  The analysis of the results of 10 runs of 10-fold cross-validation on 54 data sets (that means a total of $5400$ observations) takes about three minutes on a standard computer.

           \begin{table}[!h]
    \rowcolors{1}{white}{lightblue}
  \begin{center}
  \begin{tabular}{rrrrr}
\text{Classif. 1}  & \multicolumn{1}{c}{Classif. 2} & \text{left} &  \text{rope}  & \text{right}\\\hline
    nbc & aode &    0&	0.28&	0.72\\
nbc & hnb & 0 &	0	& 1\\
nbc & j48 & 0.2 &	0.01&	0.79 \\
nbc & j48gr & 0.15 &	0.01	& 0.84 \\
aode & hnb & 0 & 1	& 0 \\
aode & j48 & 0.46 &	0.51	&0.03 \\
aode & j48gr & 0.41 &	0.56&	0.03 \\
hnb & j48 & 0.91 &	0.07	&0.02 \\
hnb & j48gr & 0.92 &	0.05 &	0.03 \\
j48 & j48gr &0	&1	&0 \\
  \hline 
  \end{tabular}
  \end{center}
  \caption{Probabilities for the ten comparisons of classifiers.  Left and right refer to the columns Classif. 1 (left)
  and Classif. 2 (right).}
  \label{tab:hierarchicalres}
  \end{table}

\subsection{Choice of the hyper-priors parameters}
The choice of of the hyper-priors parameters can be critical in Bayesian hierarchical models.
We have conducted a sensitivity analysis by using different constants in the top-level gamma and uniform distributions, to check whether they have any notable
influence on the resulting posterior distribution.
Whether all gamma/uniform distributions are assumed, the results are essentially identical. More details about this sensitivity
analysis are reported in  \cite{corani2016unpub}. This means that for the hierarchical model inferences and decisions are stable w.r.t.\
the choice of the hyper-parameters.

\subsection{Bayesian signed rank or hierarchical model?}
\label{sec:comparison}
So far we have presented two methods for comparing two classifiers on multiple datasets:  Bayesian signed-rank and hierarchical model.
Which one should we use for comparing classifiers? In our opinion, the hierarchical model is preferable because it takes as inputs the   $m$ runs of the $k$-fold cross-validation results for each dataset and so
it makes inference about the mean difference of accuracy between two classifiers in the $i$-th dataset ($\mu_i$)  by exploiting all available information: the sample mean ($\bar{x}_i$), the variability of the data (sample standard deviation $\hat{\sigma}_i$) and the correlation due to the overlapping training set ($\rho$).
Conversely, the Bayesian signed rank only considers $\bar{x}_i$.
On the other hand, the hierarchical model is slower than  the Bayesian signed rank. In machine learning, we often need to run statistical tests
hundreds of times for instance for features selection or algorithms racing  and, in this case, it is more convenient to use a light test
as the Bayesian signed rank.

\section{Comparisons of multiple classifiers}
Another important problem with NHST is the issue of multiple hypothesis testing.
Considering the results in Table \ref{tab:pvaluefive}, which reports the $p$-values for the comparison of the five classifiers 
obtained by the Wilcoxon signed-rank test. From the  $p$-values, we concluded that ``\textit{nbc} was found significantly better than \textit{aode} and \textit{hnb},
and algorithms \textit{j48} and \textit{j48gr} were significantly different, while there were no significant differences
between other pairs''. When many tests are made,  the probability of making at least one Type 1 error in any of the
comparisons increases. One of the most popular fixes to this problem is the Bonferroni
correction. The Bonferroni correction adjusts the $p$-value at which a test is evaluated
for significance according to the number of tests being performed.
More specifically, the adjusted $p$-value is calculated as the original $p$-value divided by the number of tests being
performed. Implicitly, Bonferroni's correction assumes that these test statistics are independent. So in our current
example an overall desired significance level of $0.05$ would translate into individual tests
each using a $p$-value threshold of $0.05/10 = 0.005$ (we are performing $10$ comparisons).
In this case, this would not change our previous sections, since all significant $p$-values were less than $ 0.005$.
The Bonferroni correction reduces false rejections but it also
increases the number of instances in which the null is not rejected when actually it should have
been. Thus, the Bonferroni adjustment  can reduce the power to detect an important
effect. Motivated by this issue of the Bonferroni correction, researchers have proposed
alternative procedures. The goal of these methods typically is to reduce the family-wise error rate (that is, the probability of having at least one false positive) without 
sacrificing power too much. A natural way to achieve this is by considering  the dependence
across tests \citep{westfall1993adjusting}.

We have already discussed in Section \ref{s-freq} the pitfalls of NHST Type I error thinking.
Type~I error underlies these corrections and, therefore, corrections inherit all its problems.
The most critical one is that the correction factor depends on the way the analyst intends to conduct
the comparison. For instance, the analyst may want to compare \textit{nbc} with the other four classifiers (in this case the Bonferroni
correction would be  $0.05/4 = 0.0125$---he/she is conducting only four comparisons) or to perform all the ten comparisons ($0.05/10$) and so on. 
This creates a problem because two analysts can c draw different
conclusions from the same data because of the variety of comparisons that they made. Another important issue
with the multiple-comparison procedure based on mean-ranks test is described in \cite{benavoli2015c}.

How do we manage the problem of  multiple hypothesis testing in Bayesian analysis? Paraphrasing \cite{gelman2012we}: ``in Bayesian analysis
we usually do not have to worry about multiple comparisons.
The reason is that we do not worry about Type I error, because the null hypothesis is hardly believable to be true.''
How does Bayesian Analysis mitigate false alarms? \cite{gelman2012we} suggest
using multilevel analysis   (in our case a hierarchical Bayesian model on multiple classifiers).
Multilevel models perform partial pooling; they shift estimates toward each other.
This means that the comparisons of the classifiers are more conservative, in the sense that intervals for comparisons are more likely to include
zero. This may be a direction to pursue in future research. In this paper, we  instead mitigate false alarms 
through the rope. The rope mitigates false alarms because it decreases the asymptotic false alarm rate \citep{kruschke2013bayesian}.

\newcommand{\specialcell}[2][c]{%
  \begin{tabular}[#1]{@{}c@{}}#2\end{tabular}}
\section{Software and available Bayesian tests}
All the tests that we have presented in this paper are available in \textit{R} and \textit{Python} code at
\begin{center}
 \url{https://github.com/BayesianTestsML/tutorial/}.
 \end{center} 
Moreover,  the code that is necessary to replicate all the analyses we  performed  in this paper is 
also available at the above URL  in form of \textit{Ipython} notebooks (implemented in \textit{Python} and \textit{Julia}). 
The software is open source and that can be freely used, changed, and shared (in modified or unmodified form).

Machine learning researchers may  be interested in using other Bayesian tests besides the ones
we have discussed in this paper.
General Bayesian parametric tests can be found in \cite{kruschke2015doing} (together with R code)
and also in \cite{gelman2013bayesian}.
We have specialized some of these tests to the case of correlated data, such as the Bayesian correlated t-test \citep{coraniML2015}
discussed in Section \ref{sec:student}.
We have also implemented several Bayesian nonparametric  tests for comparing algorithms: Bayesian rank test \citep{benavoli2014b}, Friedman test \citep{benavoli2015b} and tests that account for censored data \citep{mangili2015a}.
Finally, we have developed an extension of the Bayesian sign test to compare algorithms taking
into account multiple measures  at the same time (accuracy and computational time for instance) \citep{Benavoli2015}.
For the analysis of multiple data sets, another approach has been proposed by \cite{lacoste2012bayesian} that models each data set as an
independent Bernoulli trial. The two possible outcomes of the Bernoulli trial are the first classifier
being more accurate than the second or vice versa. This approach yields the posterior probability
of the first classifier being more accurate than the second classifier on more than half of the $q$
data sets. A shortcoming is that its conclusions apply only to the q available data sets without
generalizing to the whole population of data sets

 \section{Conclusions}
 We discourage the use of frequentist null hypothesis significance tests (NHST)
 in machine learning and,  in particular, for comparison of the performance of classifiers.
 In this, we follow the current trends in other scientific areas. For instance, the journal of Basic and Applied Social Psychology, has banned the use of NHSTs and related statistical procedures~\citep{banned}. 
We believe that also in machine learning is time to move on from NHST and $p$-values. 

 In this paper,  we have discussed how Bayesian analysis can be employed instead of  NHST.
 In particular, we have presented three Bayesian tests: Bayesian correlated t-test, Bayesian signed rank test and a Bayesian hierarchical model
 that can be used for comparing the performance of classifiers and that solve the drawbacks of the frequentist tests. 
 All the code of these tests is freely available and so researchers can already use these tests for their analysis.
 In this paper, we have mainly discussed the use of Bayesian tests for comparing the performance of algorithms.
 However, in machine learning, NHST statistical tests are also employed inside the algorithms.
 For instance, nonparametric tests are used in racing algorithms, independence tests are used to learn the structure of Bayesian networks, etcetera.
 Bayesian tests can be used to replace all NHST tests because of their advantages (for instance, Bayesian tests can assess whether two algorithms are similar through the use of the rope \citep{benavoli2015b}).
  
  \section*{Acknowledgements}
Research partially supported by the Swiss NSF grant no.~IZKSZ2\_162188.  
  
  \bibliography{biblio}
  \end{document}